\documentclass[runningheads]{llncs}

\def\methodName{PolyLayout}
\def\backbone{DINOv2}

\newcommand{\xx}{\boldsymbol{x}}
\newcommand{\II}{\mathrm{\bf I}}
\newcommand{\FF}{\mathrm{\bf F}}
\newcommand{\CF}{\mathrm{\bf C_F}}
\newcommand{\EE}{\mathrm{\bf E}}
\newcommand{\CE}{\mathrm{\bf C_E}}

\usepackage{xcolor}

\definecolor{PredColor}{RGB}{60,180,75}
\definecolor{GTColor}{RGB}{0,130,200}

\definecolor{RedColor}{RGB}{230,25,75}
\definecolor{GreenColor}{RGB}{60,180,75}
\definecolor{YellowColor}{RGB}{255,225,25}
\definecolor{OrangeColor}{RGB}{245,130,48}

\usepackage{algorithm}
\usepackage{algpseudocode}
\usepackage{bm}
\usepackage{gensymb}
\usepackage{multirow}
\usepackage{threeparttable}
\usepackage{pgfplots}
\pgfplotsset{compat=1.18}
\usepackage{pgf-pie}
\usepackage{tikz}
\usepackage{overpic}
\usetikzlibrary{shapes.geometric, arrows.meta, positioning, calc}
\usepackage{arydshln}

\usepackage{eccv}

\usepackage{eccvabbrv}

\usepackage{graphicx}
\usepackage{booktabs}

\usepackage[accsupp]{axessibility}  % Improves PDF readability for those with disabilities.

\usepackage{hyperref}

\usepackage{orcidlink}

\begin{document}

\title{\methodName{}: Multi-room Manhattan Layout Estimation} 


% Include the authors' ORCID for the camera-ready version, if at all possible.
\author{Gustav Hanning\inst{1} \and
Shaohui Liu\inst{2} \and
Rémi Pautrat\inst{3} \and
Marc Pollefeys\inst{2,3} \and
Kalle Åström\inst{1} \and
Viktor Larsson\inst{1}}

\authorrunning{Hanning et al.}

\def\instnoskip#1{$^{#1}$}
\newcommand{\samelineand}{\qquad}
\institute{\instnoskip{1}Lund University
\samelineand \instnoskip{2}ETH Zurich
\samelineand \instnoskip{3}Microsoft Spatial AI Lab}

\newbool{doSupplementary}
\setbool{doSupplementary}{false}

\ifbool{doSupplementary}{
\setcounter{figure}{5}
\setcounter{section}{6}
\setcounter{table}{4}
\def\maketitlesupplementary
   {
   \newpage
   {
        \centering
        \Large
        \textbf{\methodName{}: Multi-room Manhattan Layout Estimation}\\
        \vspace{0.5em}Supplementary Material \\
        \vspace{1.0em}
    }
}

\clearpage
\setcounter{page}{1}
\maketitlesupplementary

\section{Experimental Setup}
\label{sec:supp-experimental-setup}

\subsection{Implementation Details}
\label{subsec:supp-implementation-details}

When initializing the room layout polygon $\mathcal{P}$ we estimate the orientation $\bm{R}$ by taking five LM steps to minimize $E_{VP}(\mathcal{P})$, with $\tau$ starting at 1 and decaying by half after each iteration. The polygon buffer distance $\delta$ is set to 3 m and we perform the Manhattan rasterization with a pixel size of 1 m.

The layout is simplified after each optimization step. Only walls that have converged (defined as $|\Delta d_i| < 5$ cm) can be removed. After optimization on the coarse and medium scales the polygon is split: we iteratively split the widest wall until all walls are narrower than 1 m or a maximum number of planes $p_{max} = 32$ is reached.

To create the probability map $\CF^\kappa$ for point sampling we set $\kappa = 4$. The perimeter cost $E_{per}$ is disabled ($\gamma= 0$) at the finest scale to avoid biasing the final layouts.
We resize the input images $\{ \II_i \}_{i=1}^n$ to 512 pixels in the minimum dimension.
%For ASE we convert them to grayscale as we found that it improved results.

\subsection{Network Architecture}
\label{subsec:supp-network-architecture}

Our neural network is inspired by MoGe-2 \cite{wang2025moge2}, having a \backbone{} \cite{oquab2023dinov2} ViT encoder and two convolutional decoders (heads). One decoder outputs the feature map $\FF$ and its associated confidence map $\CF$. The other predicts the edge map $\EE$ along with confidence $\CE$. Unlike MoGe-2 the two heads do not have a shared neck. We use a pre-trained ViT-S/14 model without registers for the encoder.

During training the weights of the encoder are not frozen,
%but in the second training stage
but we utilize a lower learning rate. For both training and inference we set a target number of tokens (1225) from which a patch-level resolution $h \times w$ is computed. Input images are then resized to $14h \times 14w$, with $14$ being the patch size.

The convolutional heads follow the design of MoGe-2 and progressively upsample the images from $h \times w$ to $16h \times 16w$.
%As in their work we inject a UV positional encoding at each scale.
The images in the resulting pyramid are resized with bilinear interpolation to match the input image size at the corresponding scales. For our multi-scale optimization we use the outputs at the 1/16, 1/4 and 1/1 scales. The number of channels in the feature maps are 128, 128 and 32, respectively.
We apply layer normalization \cite{ba2016layer} to the input of the residual blocks in the convolutional decoders and group normalization \cite{wu2018group} to their hidden layers.

\subsection{Training Details}
\label{subsec:supp-training-details}

The network is trained as in \cite{hanning2025pixcuboid}, but our training data consists of a mix of cuboid and non-cuboid room layouts. We sample four times more training examples per room from our 107 manually annotated Manhattan layouts than from the 391 cuboid scenes in their training set.

We noticed that the pre-training converged much faster with \backbone{}, so the number of epochs is decreased from 10 to 3. In the second stage of training we use a batch size of 2 and set the initial learning rate to $3.33 \times 10^{-7}$ for the encoder and $3.33 \times 10^{-6}$ for the convolutional heads. After 10 epochs it is reduced by a factor of 5 and training continues for another 5 epochs. The two training stages take around 4 h and 72 h, respectively, on one NVIDIA TITAN V GPU with 12 GB of memory. 

%The network is trained as in \cite{hanning2025pixcuboid}, except for the change in epochs and learning rate schedule described in \cref{subsec:training}. Additionally, our training data consists of a mix of cuboid and non-cuboid room layouts. We sample four times more training examples from the manually annotated scenes with Manhattan layouts than from the cuboid scenes in their training set.

\subsection{Image Sampling}
\label{subsec:supp-image-sampling}

\textbf{Aria Synthetic Environments}: As the images originate from a semi-dense camera trajectory there are typically many close-by cameras with similar viewing direction. We noted that sampling images uniformly often resulted in image sets with poor coverage of the scene and walls that were not visible in any view. For that reason we employ a visibility-based sampling scheme that tries to maximize the shared visual coverage. A grid of 3D points, 0.25 m in between, is generated for the floor, ceiling and walls of a room. Images are selected iteratively with a probability proportional to
\begin{equation}
    p = \sum_{i=1}^N v_i |\mathbf{n}_i \cdot \mathbf{d}_i| \eta^{-k_i},
\end{equation}
where $N$ is the total number of points. $v_i$ is one if the point is visible in the image and zero otherwise. $\bm{n}_i$ is the normal vector of the surface and $\bm{d}_i$ the normalized viewing direction (vector between the camera center and 3D point). $k_i$ is the number of images previously selected where the point $i$ is visible and $\eta$ a scale factor which we set to 10. Images that are outside the ground truth room layout are assigned zero probability.
This simple heuristic promotes sampling images that see previously unobserved points while taking the angle the wall is seen from into account.

Prior to sampling we filter out images that are inside one room but capture the contents of another room. This is done by rendering the doors leading to adjacent rooms into the view and counting the number of pixels. If more than 10\% of the image is filled with such door pixels it is excluded from sampling.

%Even with the visibilty-based sampling and exclusion of images "looking into" other rooms this remains a challenging dataset. We tune our method on the validation set but do not use it for training.

\textbf{ScanNet++}: As ScanNet++ has a more uniform distribution of cameras in terms of position and viewing direction than Aria Synthetic Environments we sample the images with equal probability, but similarly exclude images for which the camera center is not inside the ground truth layout. Doors are not explicitly available so instead we subtract the depth of the layout $D_{layout}$ from that of the provided semantic mesh ($D_{mesh}$) and skip images where $D_{mesh} - D_{layout} > 0.5$ m for more than 10\% of the pixels.

\section{Additional Results}

In \cref{fig:scannetpp-layouts} and \cref{fig:2d3ds-layouts} we show additional examples of predicted room layouts for \methodName{} and competing methods.
Failure cases of our method can be found in \cref{fig:failure-cases}.
We see that \methodName{} can struggle to identify all walls in rooms with very complex shapes.
This can be due to certain parts of the scene not being observed in any view, even with our visibility-based image sampling (\cref{subsec:supp-image-sampling}).
%, despite the visibility-based sampling.
\cref{fig:dino-vs-resnet} contains examples of predicted feature, edge and confidence maps for one image from Aria Synthetic Environments and one from ScanNet++, comparing the ResNet CNN from \cite{hanning2025pixcuboid} with our \backbone{} network. Both networks are trained on ScanNet++ but ours generalizes better to the synthetic data in ASE and is able to better predict the layout edges (fourth row).
%In \cref{tab:ablation-experiments} and \cref{tab:cuboid-experiments} we give the full set of metrics for the ablation experiments and the cuboid baseline comparison, respectively.

%In the supplementary material we also include two videos. \textit{22913.mp4} visualizes the optimization steps for a scene with three rooms in our Aria Synthetic Environments test set. Here we show the layouts in 3D as well as projected into one image for each room.
%In \textit{area\_6.mp4} \methodName{} optimizes the layouts for all the cuboid-shaped rooms in one area of 2D-3D-Semantics.
%The video \textit{45109.mp4} compares the predicted layouts of SceneScript, Plane-DUSt3R, PixCuboid and RoomFormer with that of \methodName{} for one scene in the ASE test set.

\begin{figure}
    \centering
    \begin{tikzpicture}[scale=1.0]
        % \node at (0.0,0.0) {\includegraphics[scale=0.08]{images/8be0cd3817_0136/SceneScript.jpg}};
        % \node[align=center] at (0.0,-1.4) {SceneScript};

        % \node at (3.7,0.0) {\includegraphics[scale=0.08]{images/8be0cd3817_0136/Plane-DUSt3R.jpg}};
        % \node[align=center] at (3.7,-1.4) {Plane-DUSt3R};
        
        % \node at (0.0,-2.9) {\includegraphics[scale=0.08]{images/8be0cd3817_0136/PixCuboid.jpg}};
        % \node[align=center] at (0.0,-4.3) {PixCuboid};

        % \node at (3.7,-2.9) {\includegraphics[scale=0.08]{images/8be0cd3817_0136/RoomFormer.jpg}};
        % \node[align=center] at (3.7,-4.3) {RoomFormer};

        % \node at (7.8,-1.45) {\includegraphics[scale=0.105]{images/8be0cd3817_0136/PolyLayout.jpg}};
        % \node[align=center] at (7.8,-3.0) {\methodName{}};

        \node at (0.0,-6.5) {\includegraphics[scale=0.08]{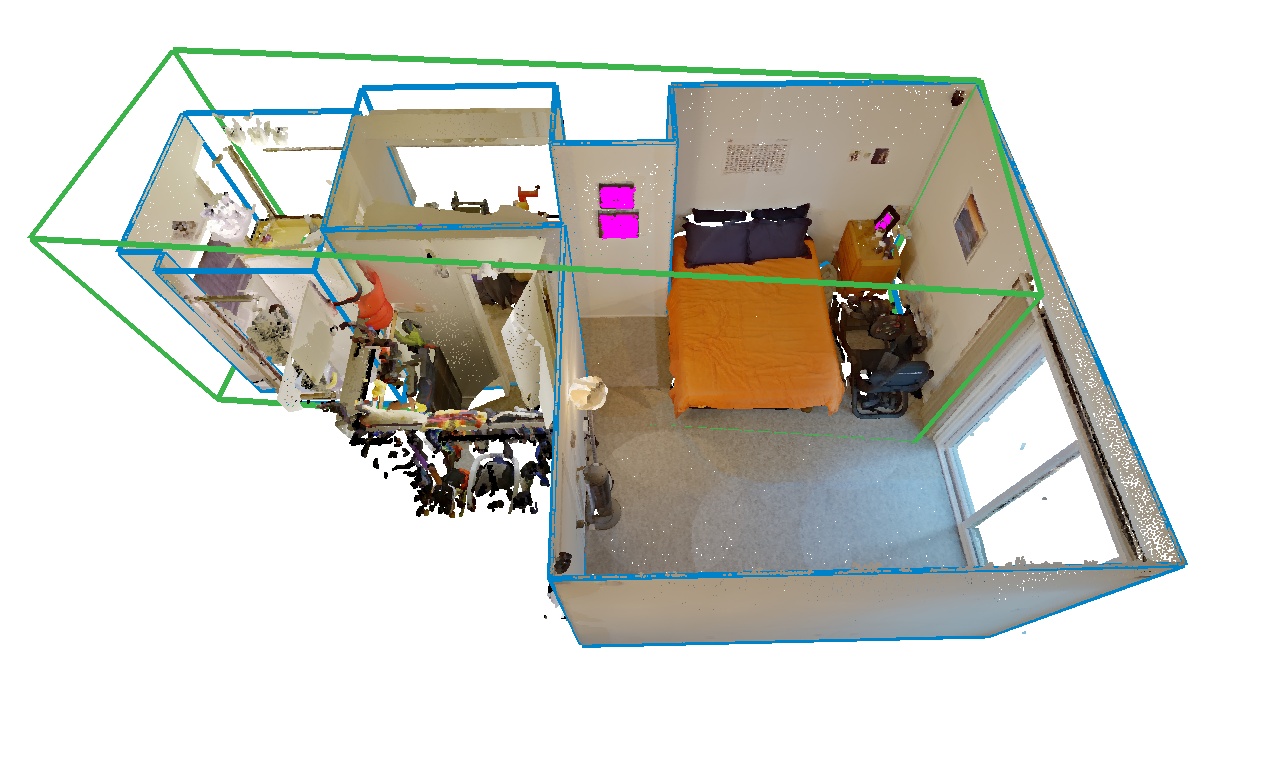}};
        \node[align=center] at (0.0,-7.9) {SceneScript};

        \node at (3.7,-6.5) {\includegraphics[scale=0.08]{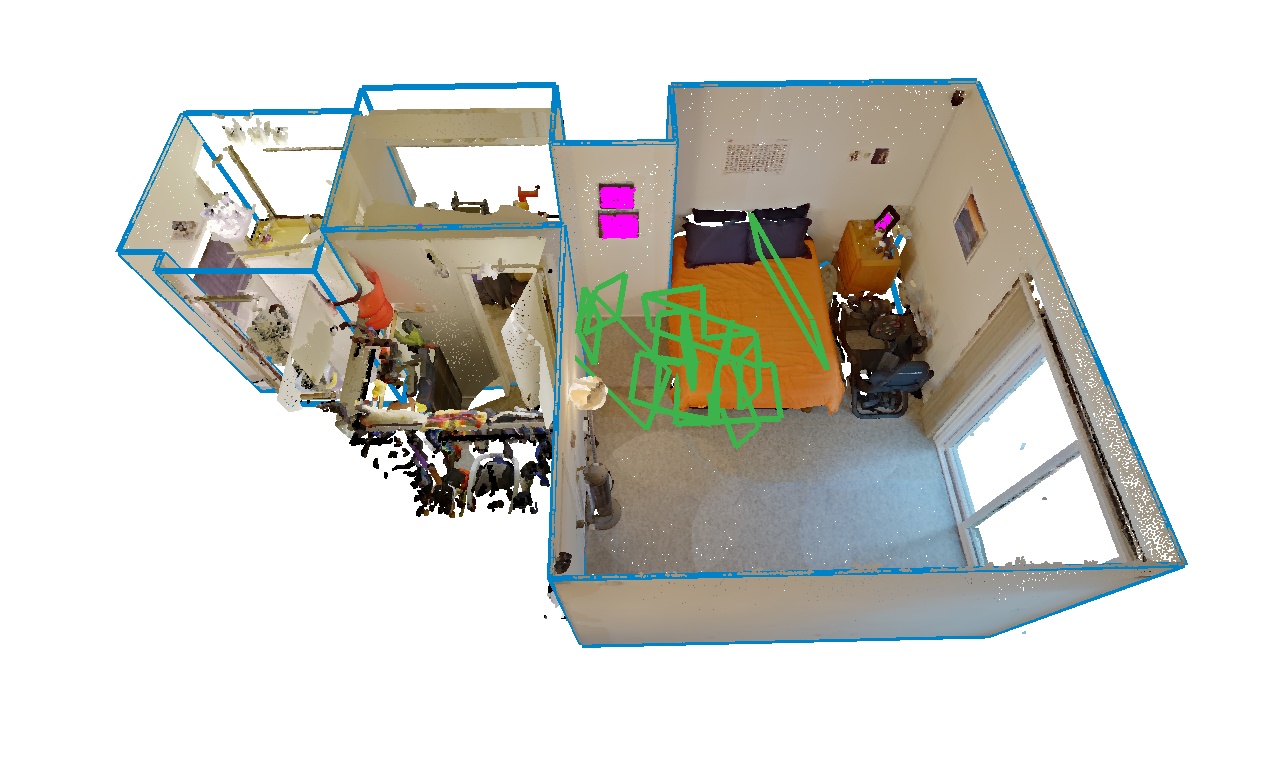}};
        \node[align=center] at (3.7,-7.9) {Plane-DUSt3R};
        
        \node at (0.0,-9.4) {\includegraphics[scale=0.08]{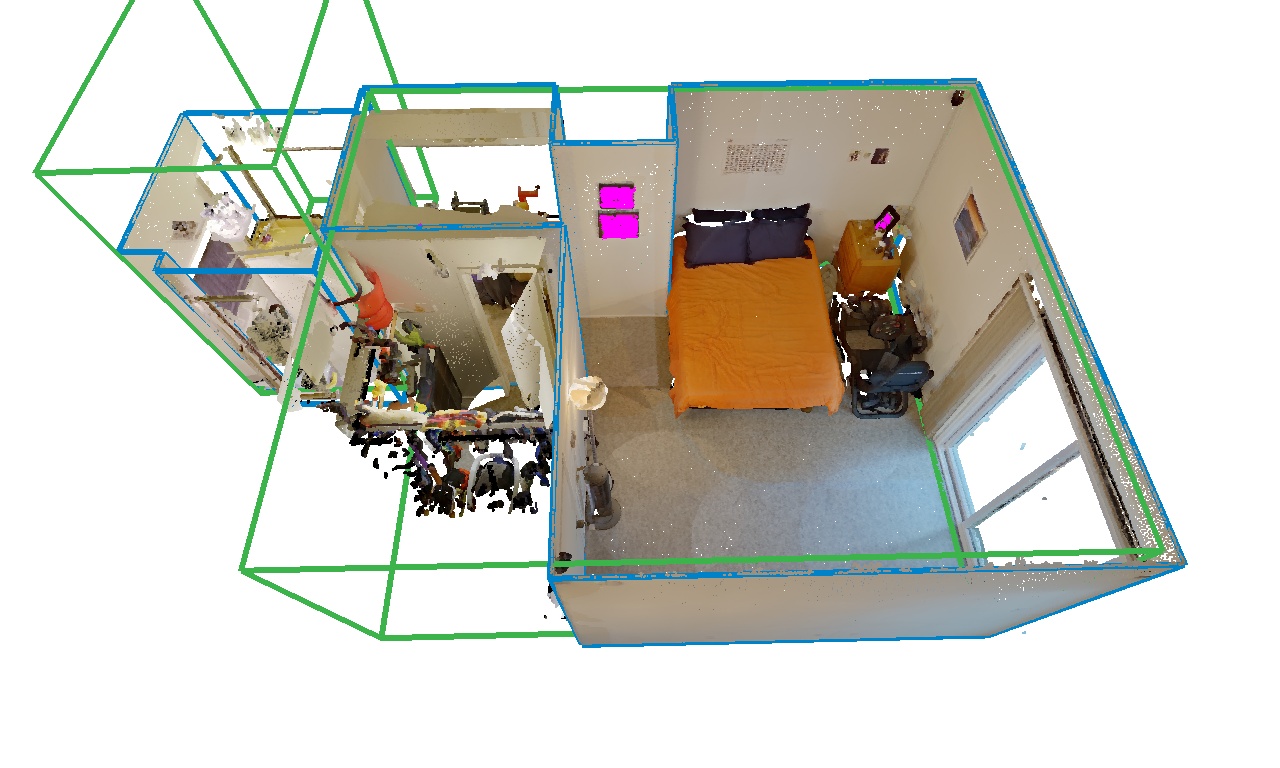}};
        \node[align=center] at (0.0,-10.8) {PixCuboid};

        \node at (3.7,-9.4) {\includegraphics[scale=0.08]{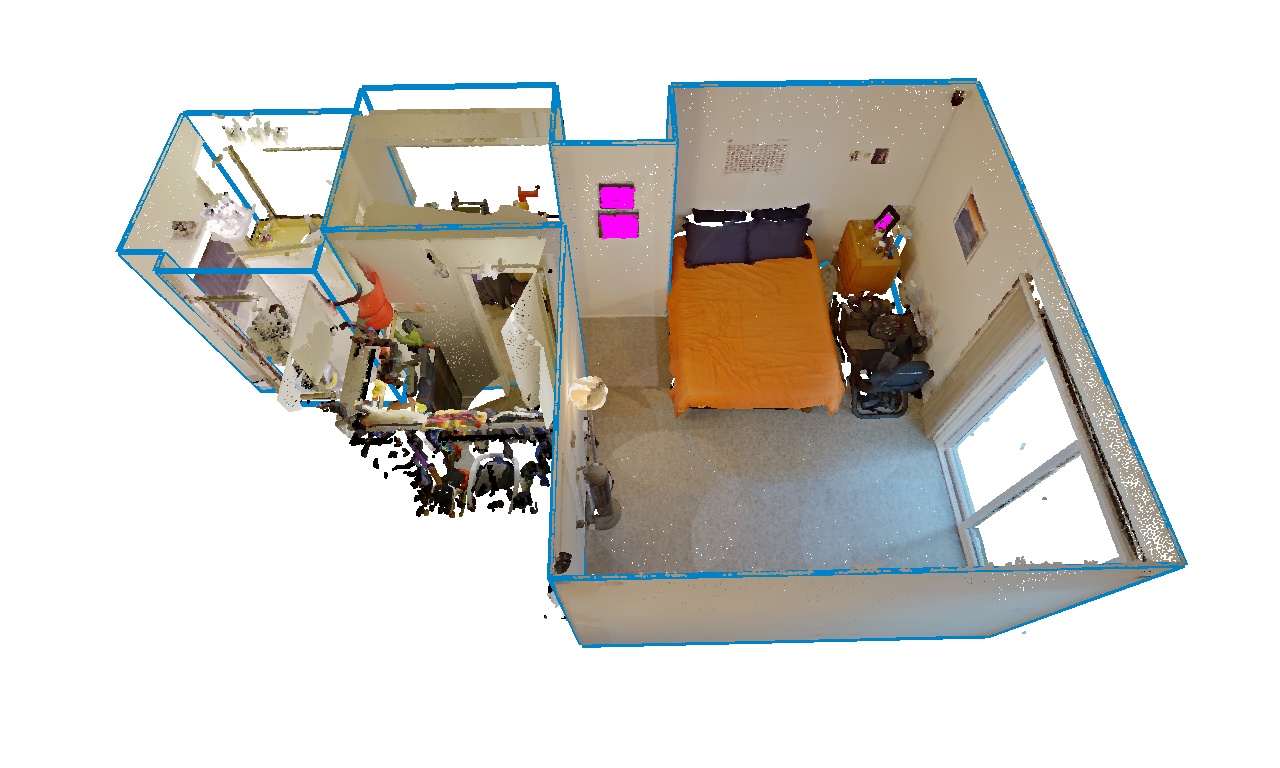}};
        \node[align=center] at (3.7,-10.8) {RoomFormer};

        \node at (7.8,-7.95) {\includegraphics[scale=0.105]{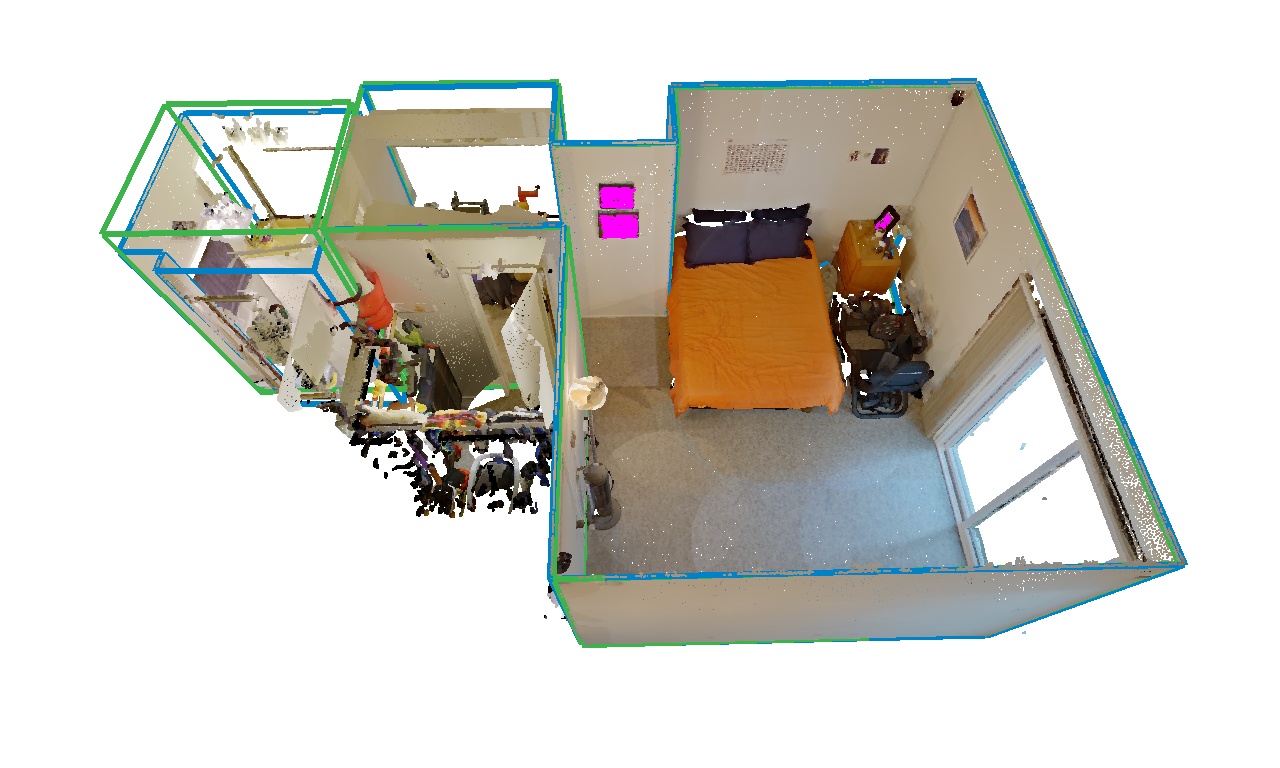}};
        \node[align=center] at (7.8,-9.5) {\methodName{}};

        %\draw[dashed] (-1.0,-5.0)--(9.0,-5.0);
    \end{tikzpicture}
    \caption{\textbf{Room layout predictions} (in \textcolor{PredColor}{\bf green}) for one scene in our ScanNet++ v2 test set. Ground truth layouts are shown in \textcolor{GTColor}{\bf blue}.}
    \label{fig:scannetpp-layouts}
\end{figure}

\begin{figure}[tp]
    \centering
    \begin{tikzpicture}[scale=1.0]
        \node at (0.0,0.0) {\includegraphics[scale=0.12]{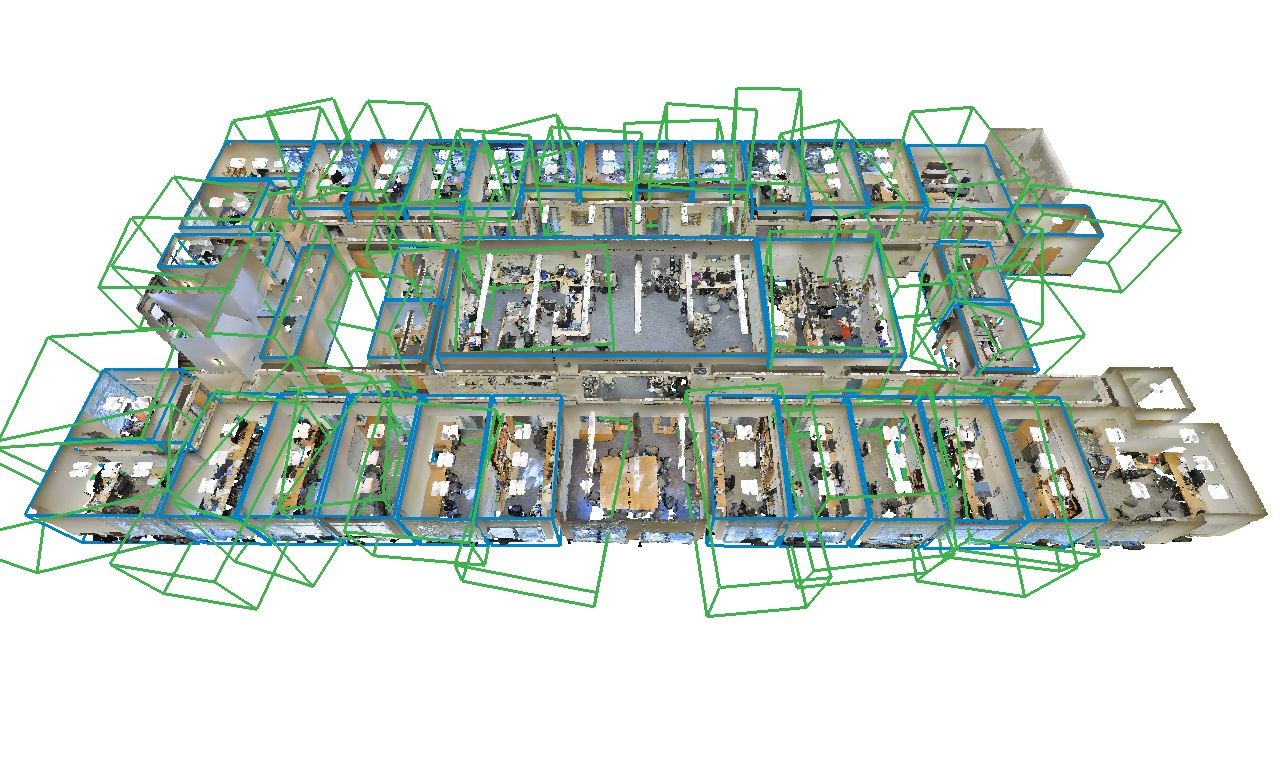}};
        \node[align=center] at (0.0,-1.3) {Total3D};

        \node at (5.7,0.0) {\includegraphics[scale=0.12]{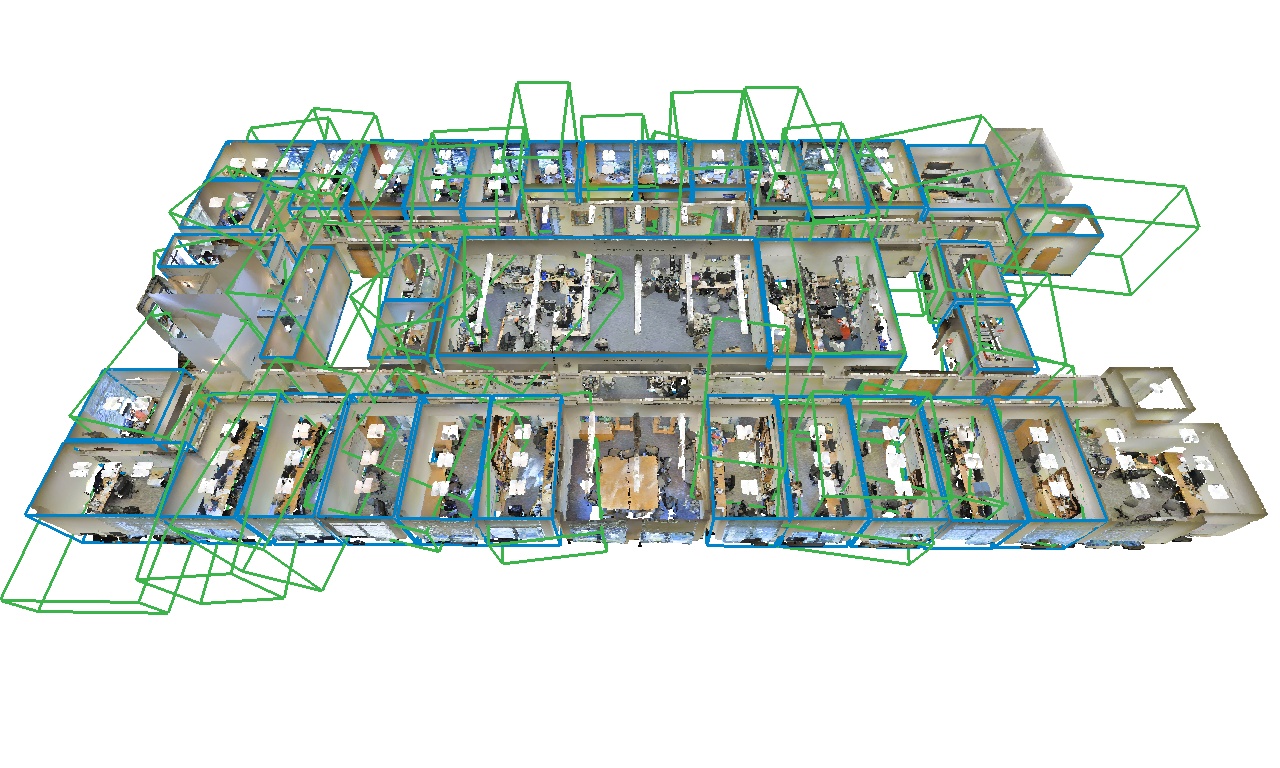}};
        \node[align=center] at (5.7,-1.3) {Implicit3D};

        \node at (0.0,-3.2) {\includegraphics[scale=0.12]{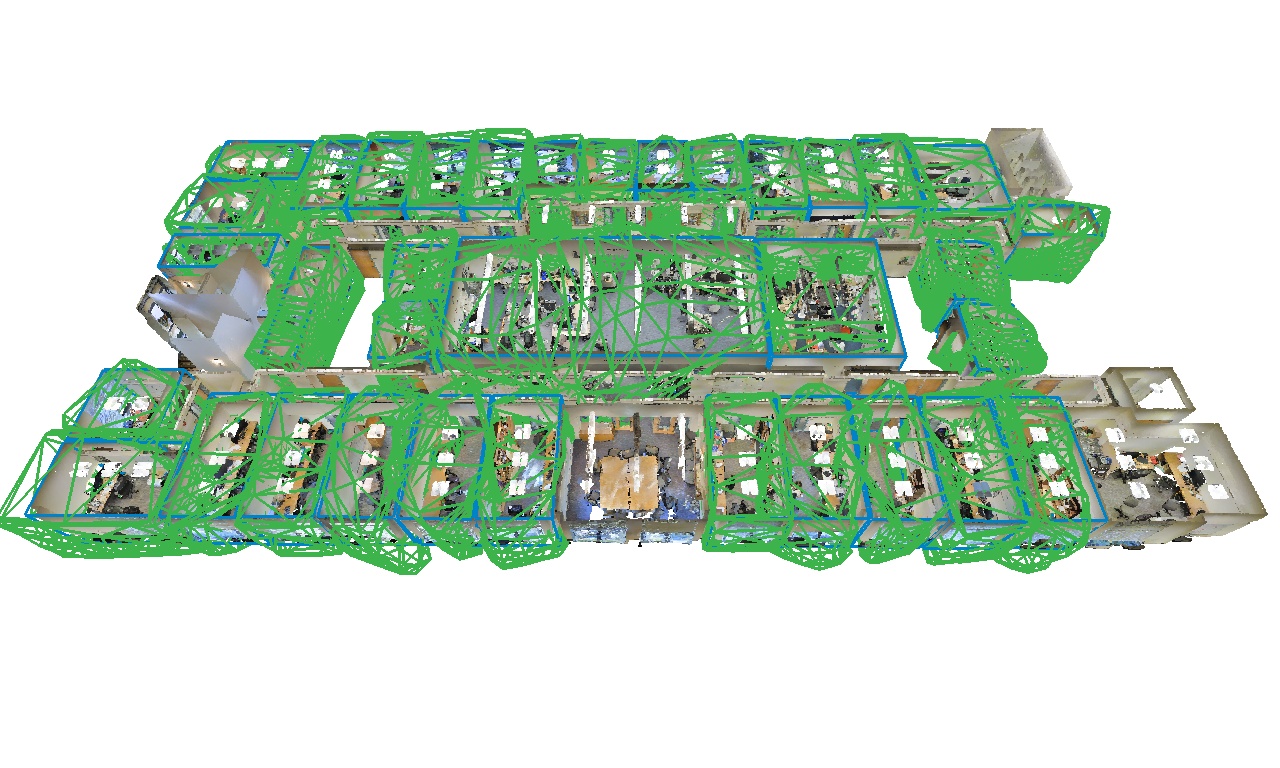}};
        \node[align=center] at (0.0,-4.5) {Deep3DLayout};

        \node at (5.7,-3.2) {\includegraphics[scale=0.12]{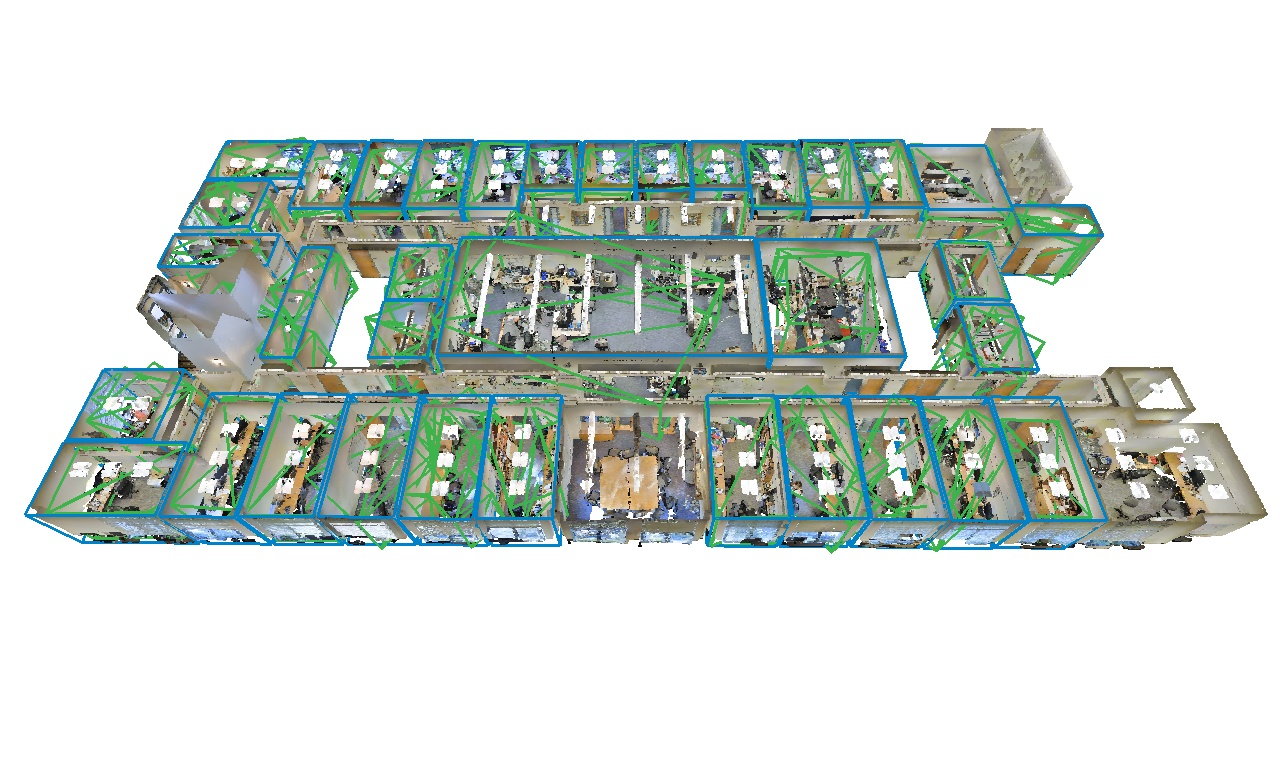}};
        \node[align=center] at (5.7,-4.5) {LED\textsuperscript{2}-Net};

        \node at (0.0,-6.4) {\includegraphics[scale=0.12]{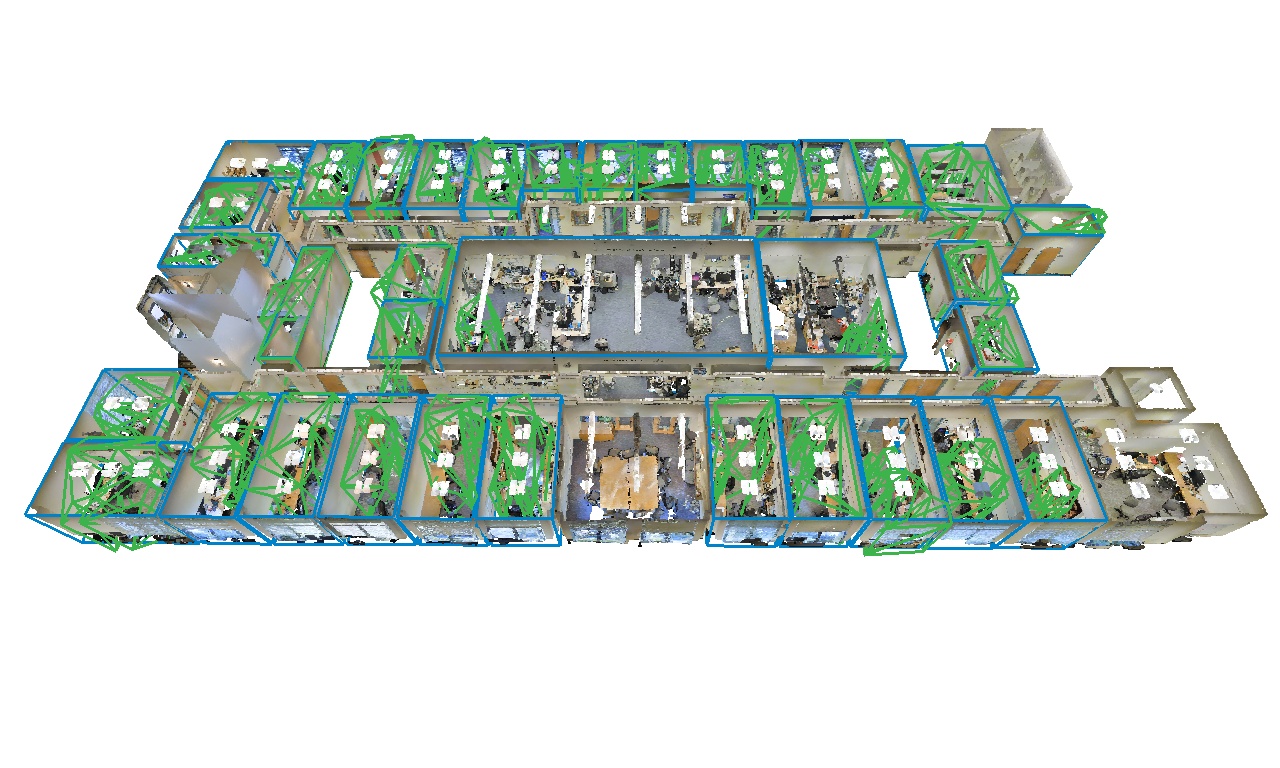}};
        \node[align=center] at (0.0,-7.7) {PSMNet};

        \node at (5.7,-6.4) {\includegraphics[scale=0.12]{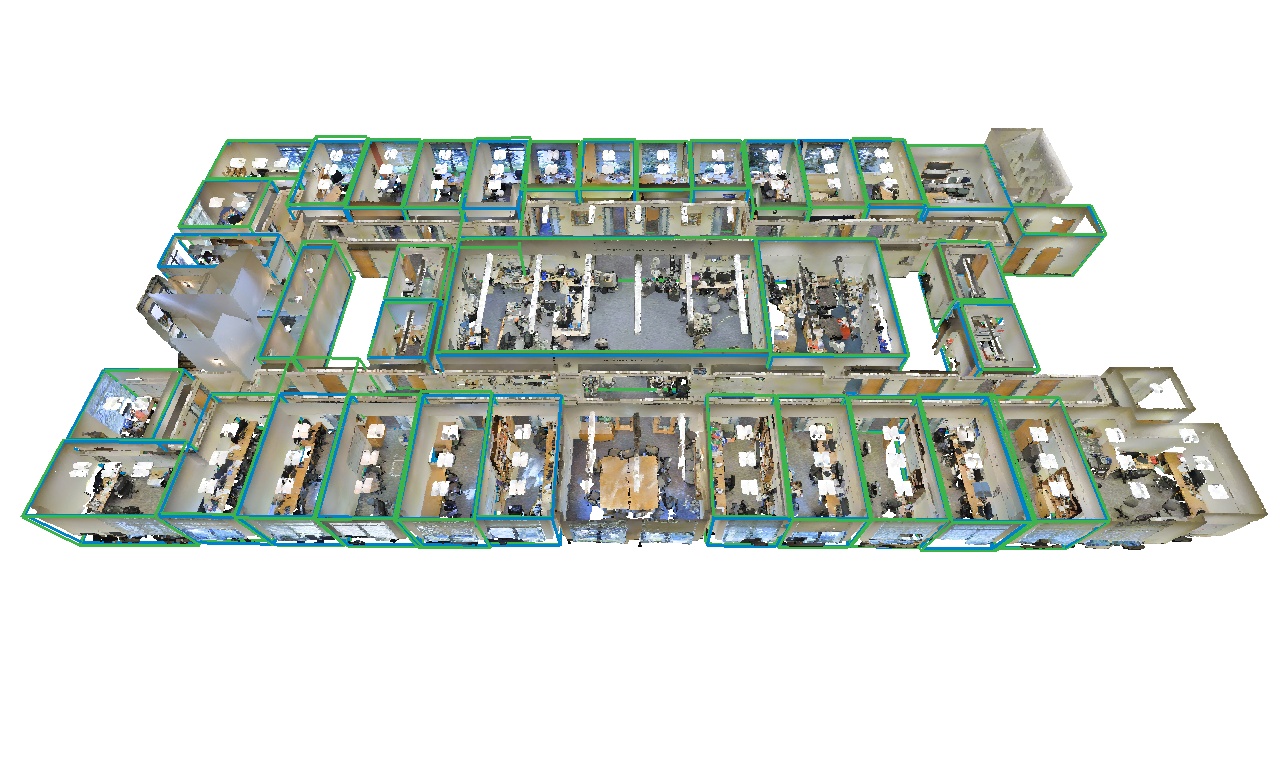}};
        \node[align=center] at (5.7,-7.7) {PixCuboid};

        \node at (2.85,-10.9) {\includegraphics[scale=0.22]{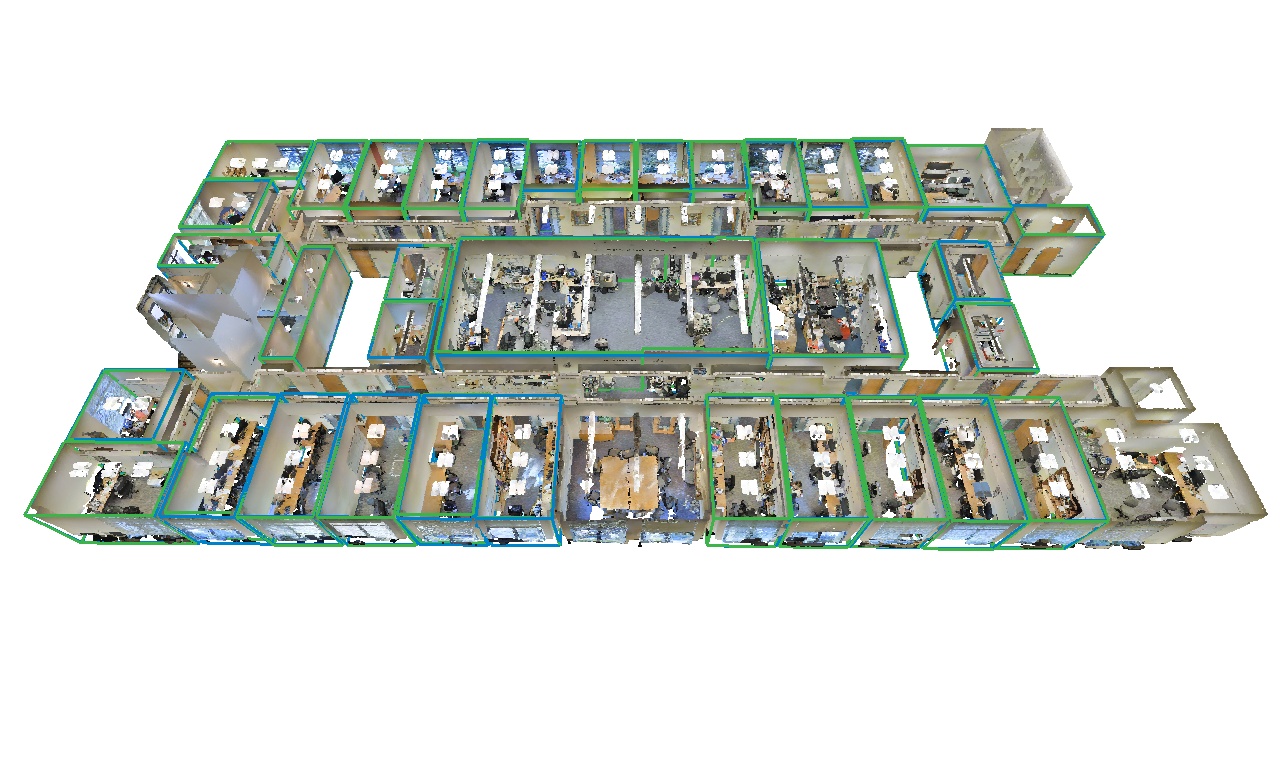}};
        \node[align=center] at (2.85,-12.8) {\methodName{}};
    \end{tikzpicture}
    \caption{\textbf{Room layout predictions} for the cuboid-shaped spaces in one area of 2D-3D-Semantics. Predictions are shown in \textcolor{PredColor}{\bf green} and the ground truth layouts in \textcolor{GTColor}{\bf blue}. For single-view methods the prediction with highest IoU is visualized.}
    \label{fig:2d3ds-layouts}
\end{figure}

\begin{figure*}[ht]
    \centering
    \begin{tikzpicture}[scale=1.0]
        \node at (0.0,0.0) {\includegraphics[scale=0.13]{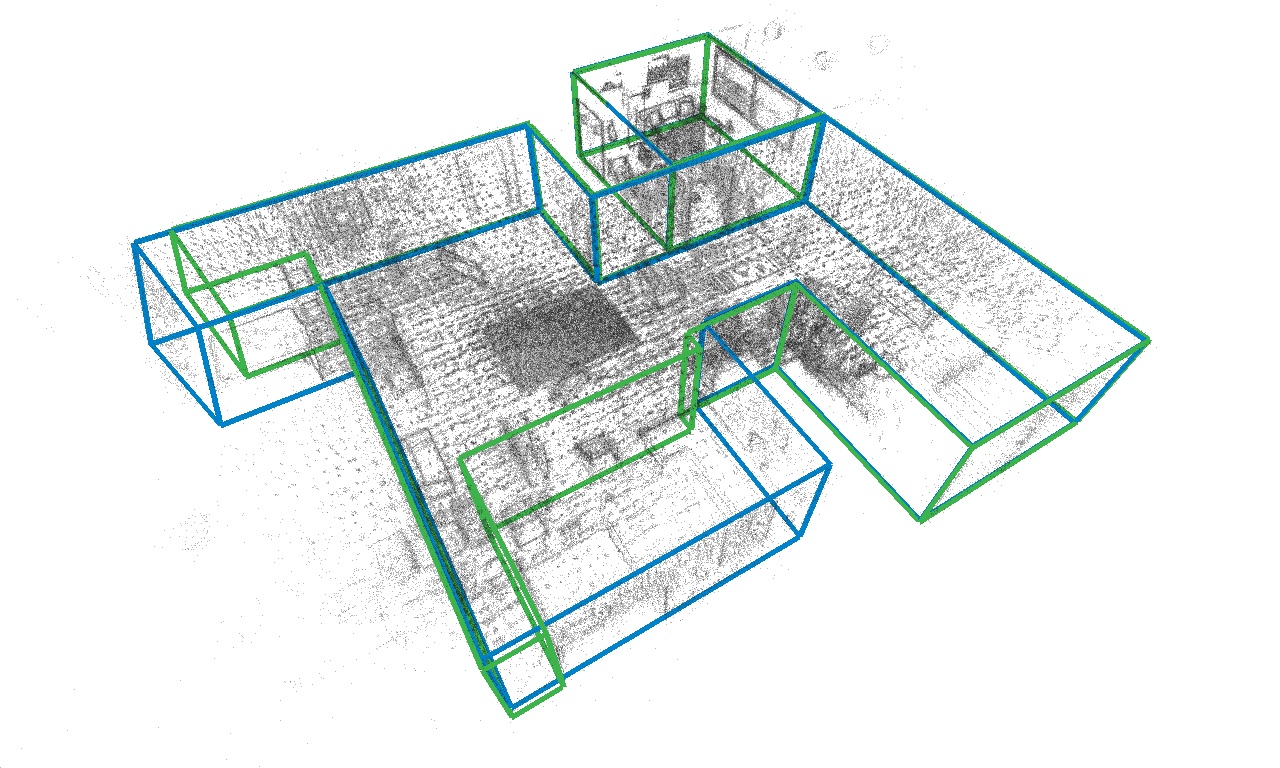}};
        \node at (5.8,0.0) {\includegraphics[scale=0.13]{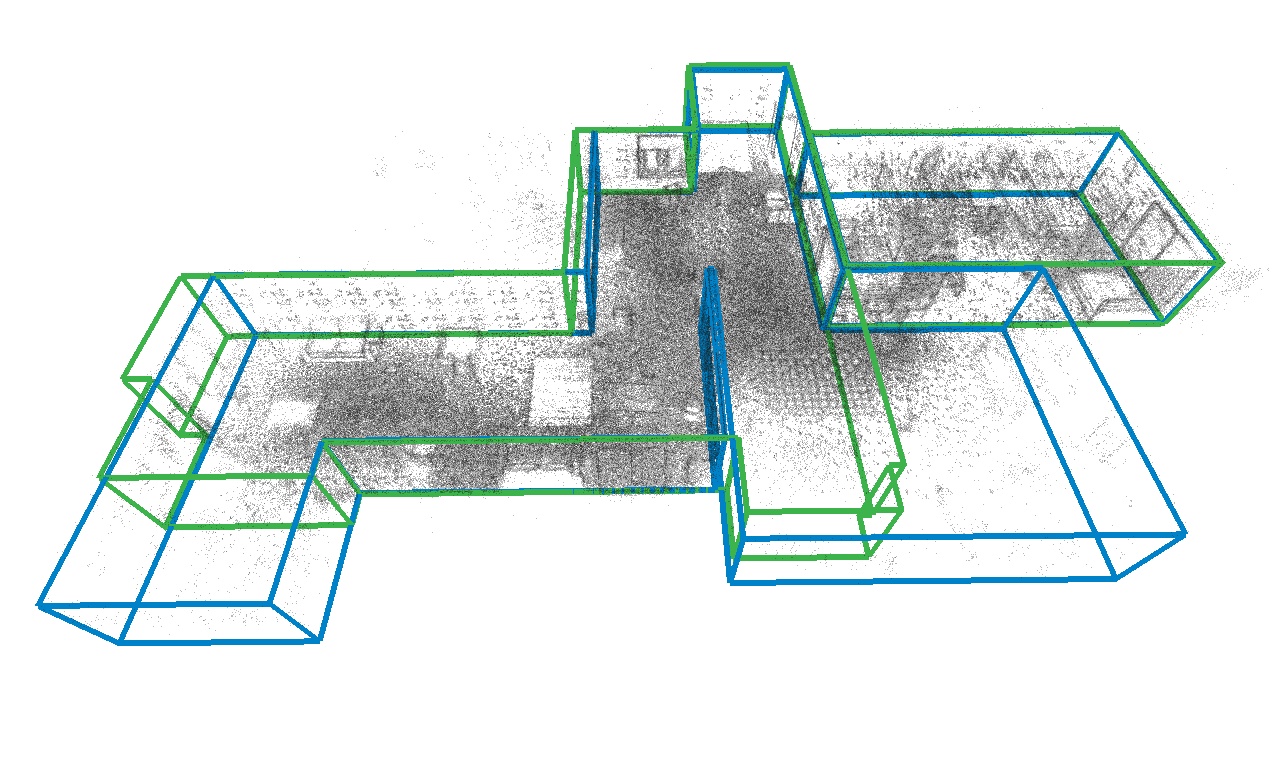}};
        \node at (0.0,4.0) {\includegraphics[scale=0.13]{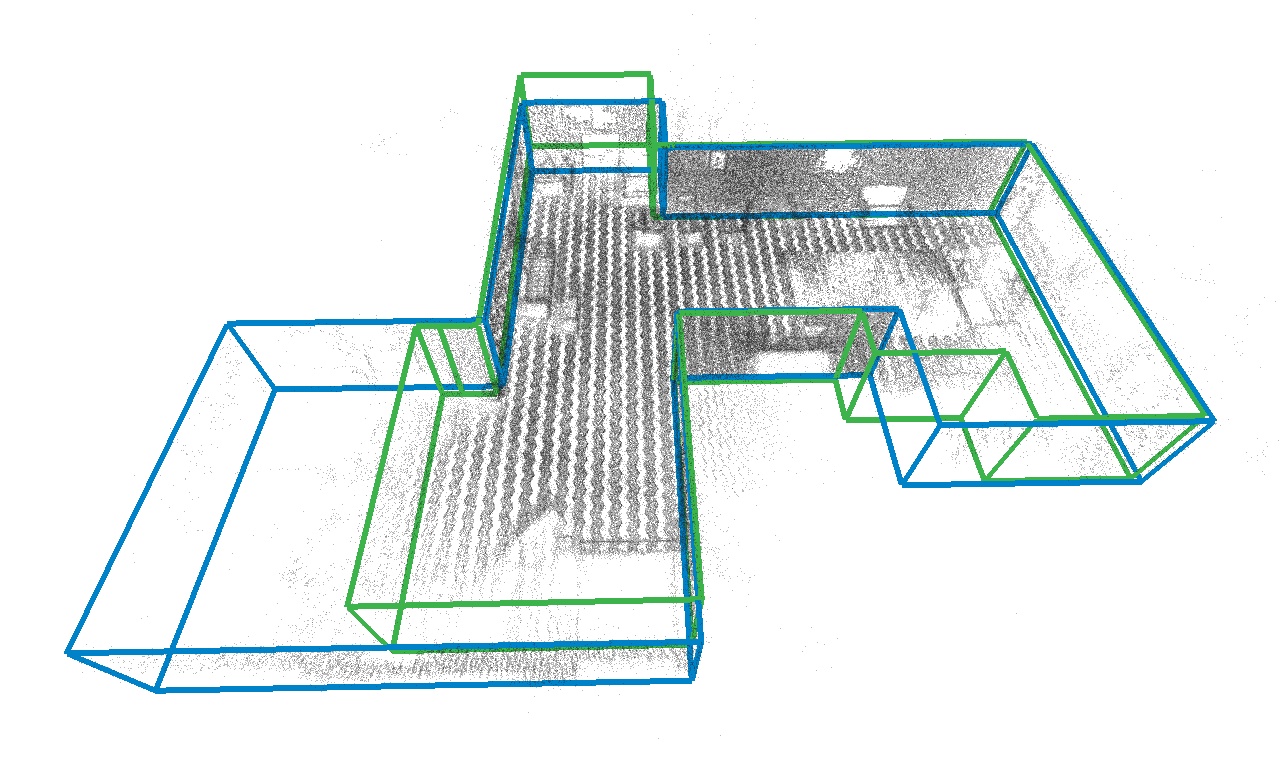}};
        \node at (5.8,4.0) {\includegraphics[scale=0.13]{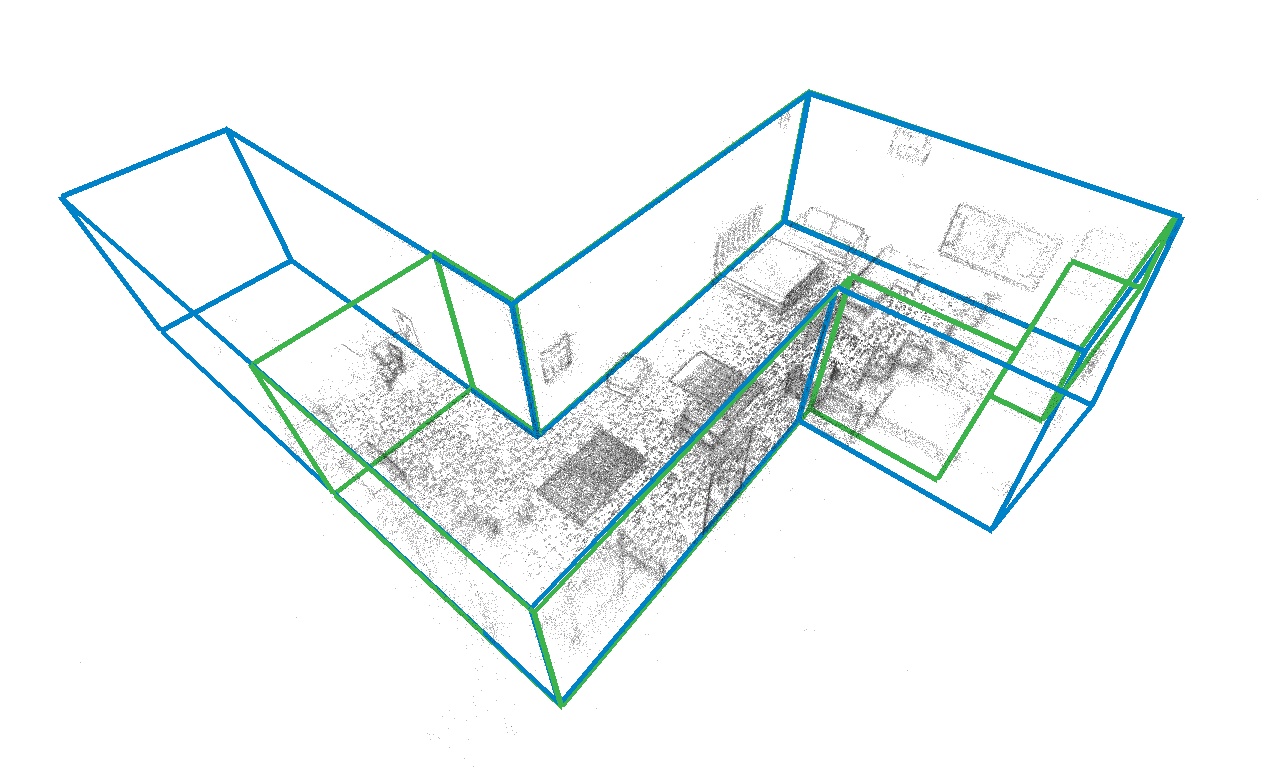}};
    \end{tikzpicture}
    \caption{\textbf{Failure cases of \methodName{}} in Aria Synthetic Environments. Predictions are shown in \textcolor{PredColor}{\bf green} and the ground truth layouts in \textcolor{GTColor}{\bf blue}.}
    \label{fig:failure-cases}
\end{figure*}

\begin{figure*}
    \centering
    \begin{tikzpicture}[scale=1.0]
        \node[align=center] at (4.15,-0.8) {\textbf{Aria Synthetic Environments}};
    
        \node[align=center] at (3.0,-1.5) {ResNet};
        \node[align=center] at (5.3,-1.5) {\backbone{}};

        \node[align=center] at (10.2,-0.8) {\textbf{ScanNet++}};
        
        \node[align=center] at (8.6,-1.5) {ResNet};
        \node[align=center] at (12.0,-1.5) {\backbone{}};

        \node[align=center,rotate=90] at (1.65,-3.0) {\scriptsize Image $\II$};
        \node[align=center,rotate=90] at (1.65,-5.3) {\scriptsize Feature map $\FF$};
        \node[align=center,rotate=90] at (1.65,-7.6) {\scriptsize Feat. conf. $\CF$};
        \node[align=center,rotate=90] at (1.65,-9.9) {\scriptsize Edge map $\EE$};
        \node[align=center,rotate=90] at (1.65,-12.2) {\scriptsize Edge conf. $\CE$};

        \node at (3.0,-3.0) {\includegraphics[scale=0.12]{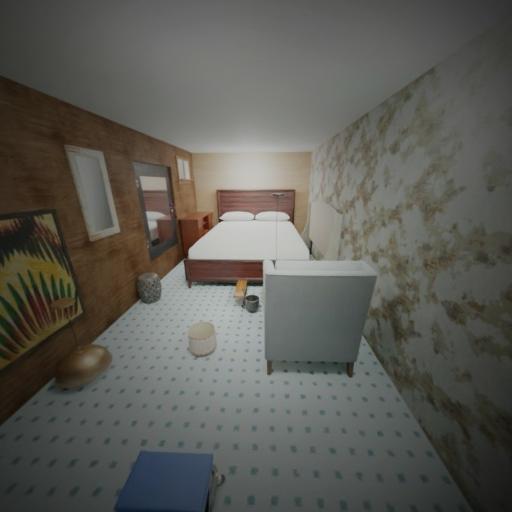}};
        \node at (3.0,-5.3) {\includegraphics[scale=0.12]{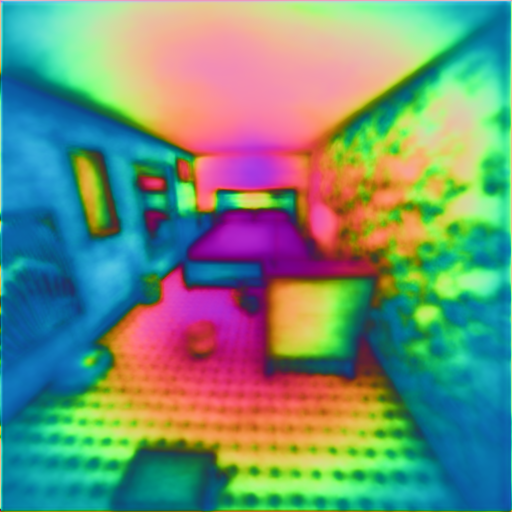}};
        \node at (3.0,-7.6) {\includegraphics[scale=0.12]{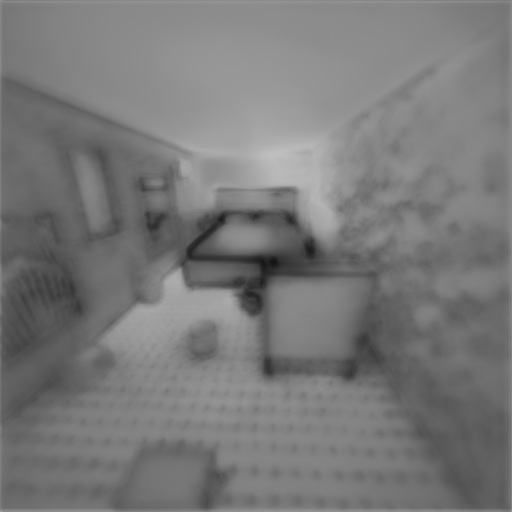}};
        \node at (3.0,-9.9) {\includegraphics[scale=0.12]{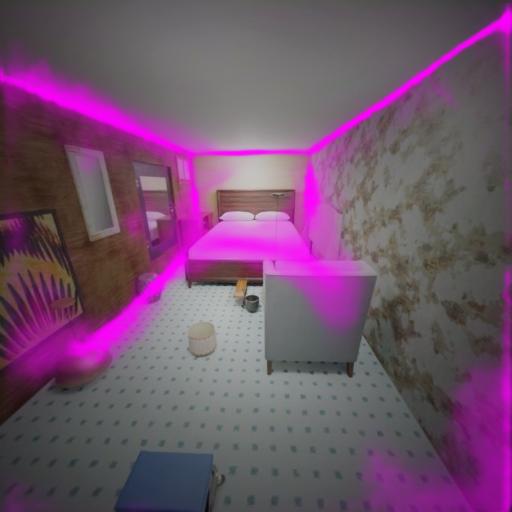}};
        \node at (3.0,-12.2) {\includegraphics[scale=0.12]{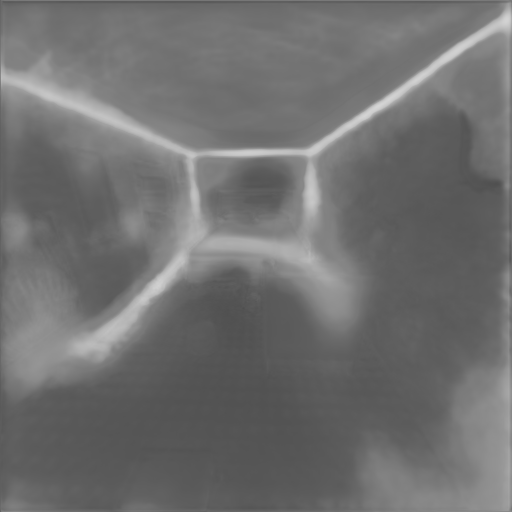}};
        
        \node at (5.3,-3.0) {\includegraphics[scale=0.12]{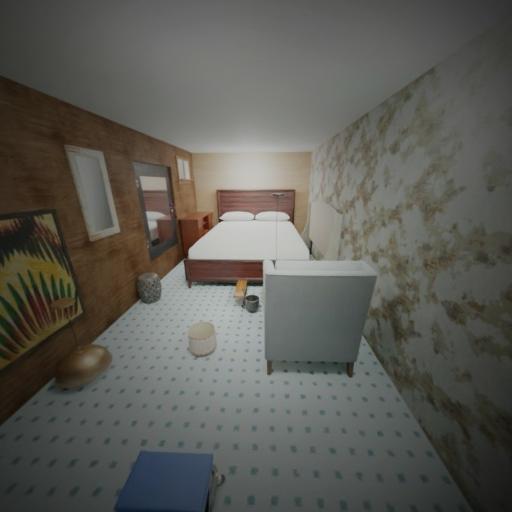}};
        \node at (5.3,-5.3) {\includegraphics[scale=0.12]{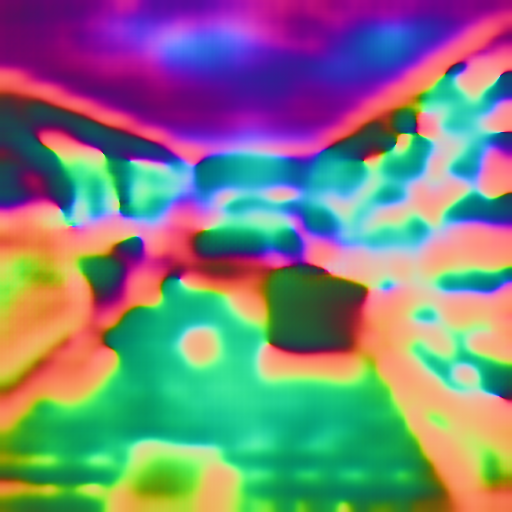}};
        \node at (5.3,-7.6) {\includegraphics[scale=0.12]{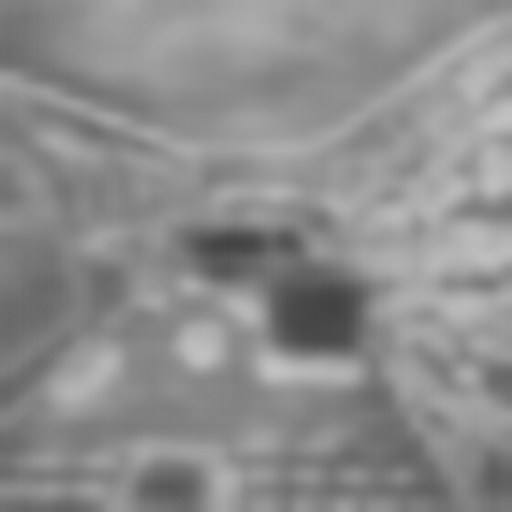}};
        \node at (5.3,-9.9) {\includegraphics[scale=0.12]{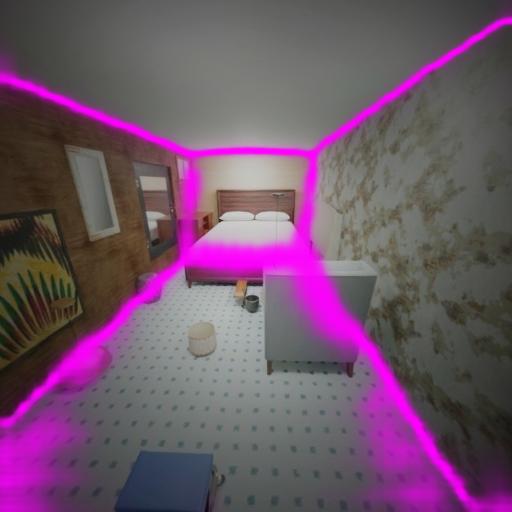}};
        \node at (5.3,-12.2) {\includegraphics[scale=0.12]{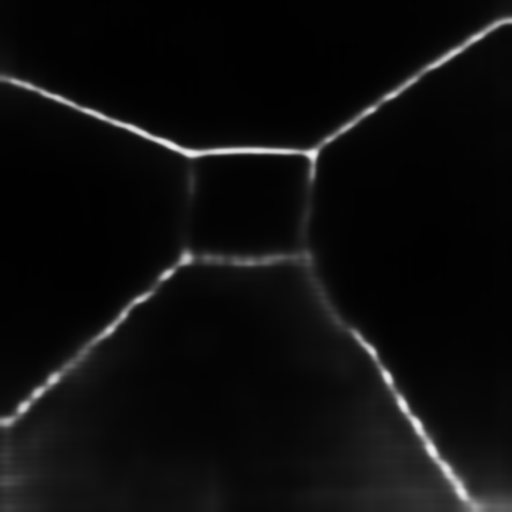}};

        \node at (8.5,-3.0) {\includegraphics[scale=0.12]{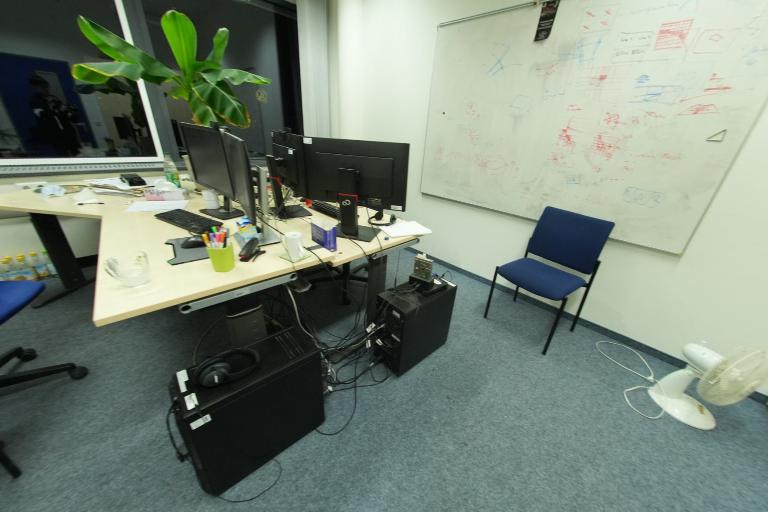}};
        \node at (8.5,-5.3) {\includegraphics[scale=0.12]{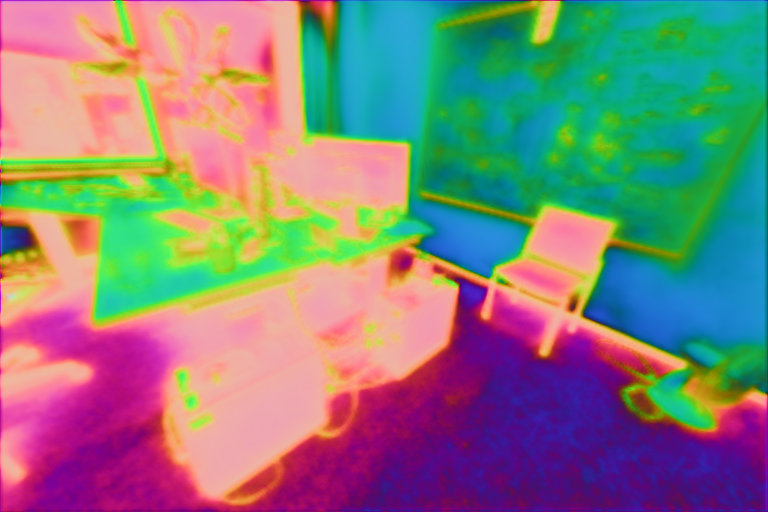}};
        \node at (8.5,-7.6) {\includegraphics[scale=0.12]{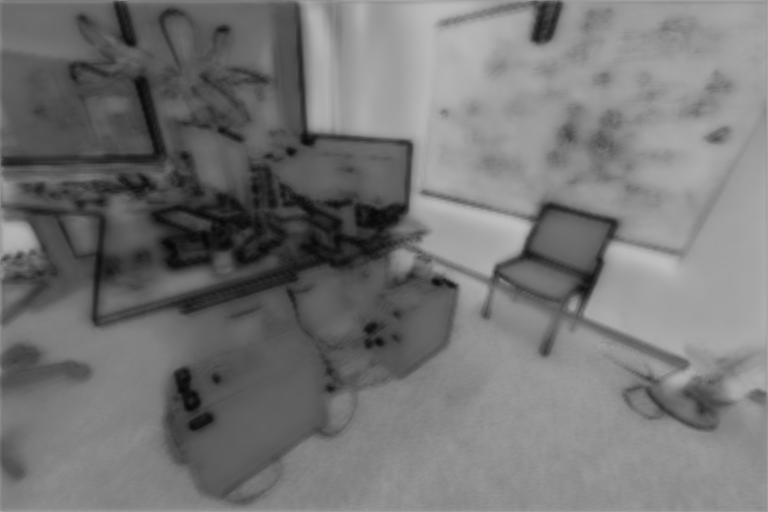}};
        \node at (8.5,-9.9) {\includegraphics[scale=0.12]{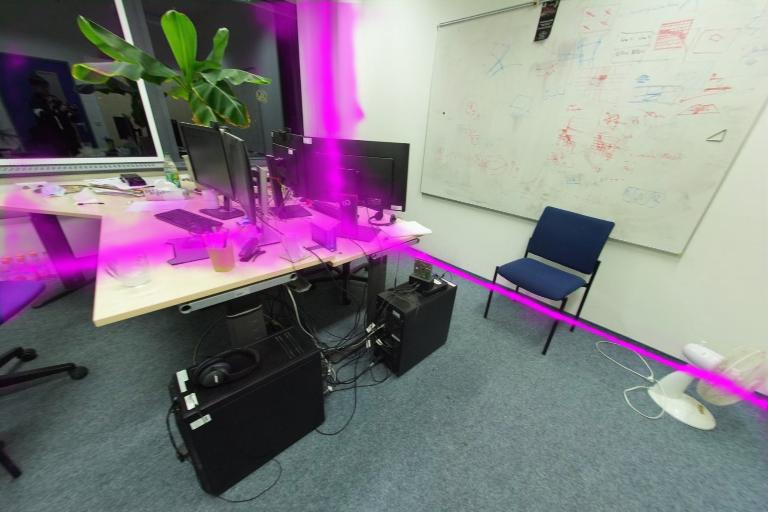}};
        \node at (8.5,-12.2) {\includegraphics[scale=0.12]{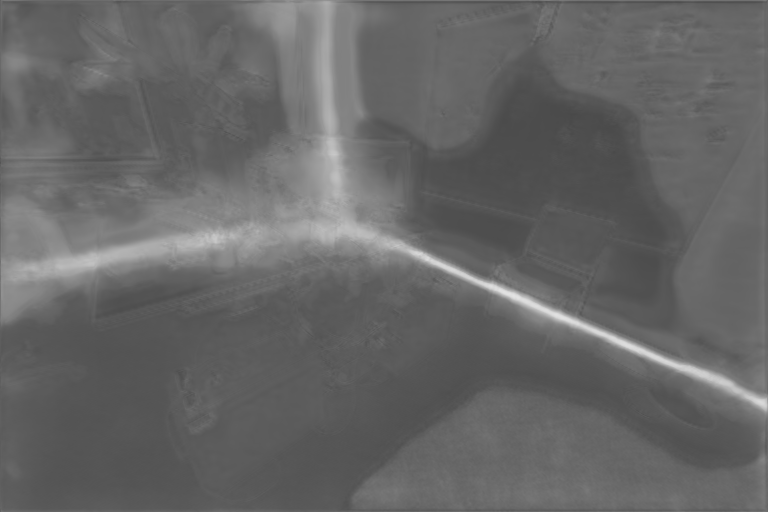}};
        
        \node at (11.9,-3.0) {\includegraphics[scale=0.12]{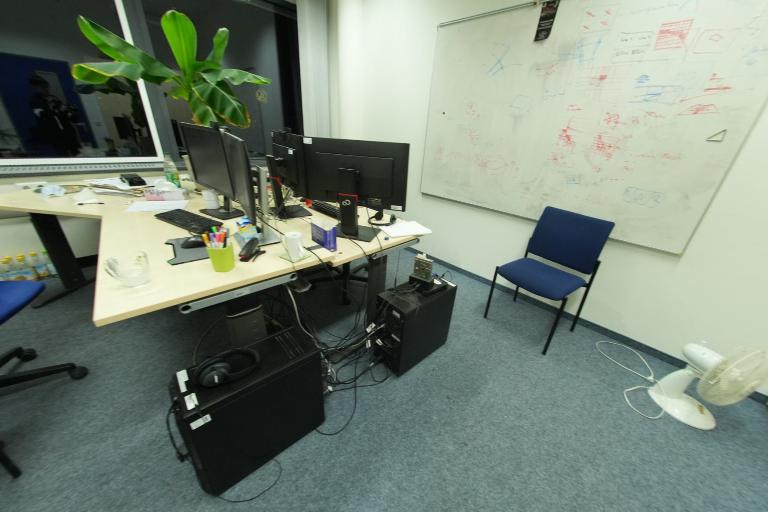}};
        \node at (11.9,-5.3) {\includegraphics[scale=0.12]{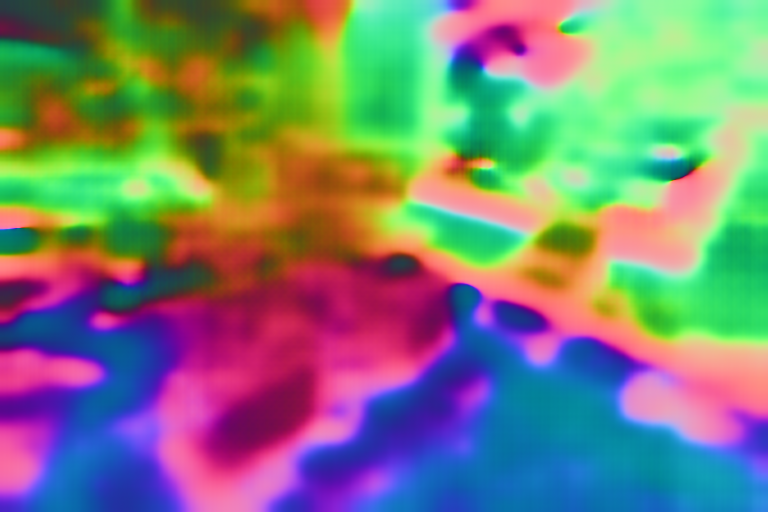}};
        \node at (11.9,-7.6) {\includegraphics[scale=0.12]{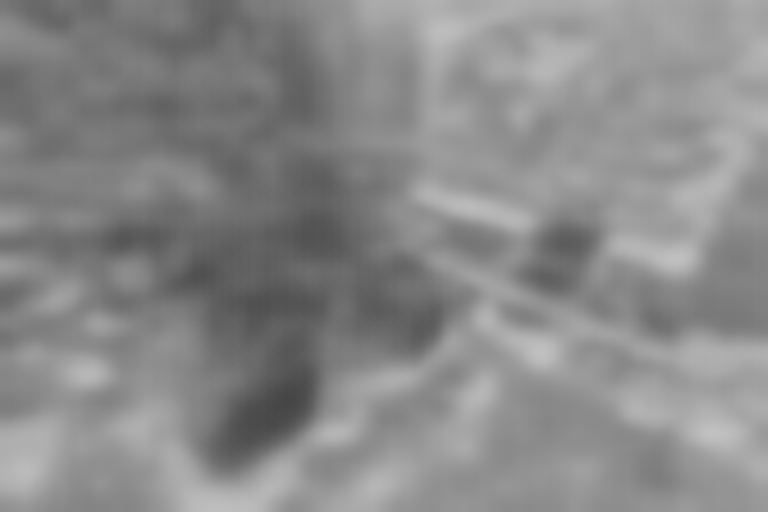}};
        \node at (11.9,-9.9) {\includegraphics[scale=0.12]{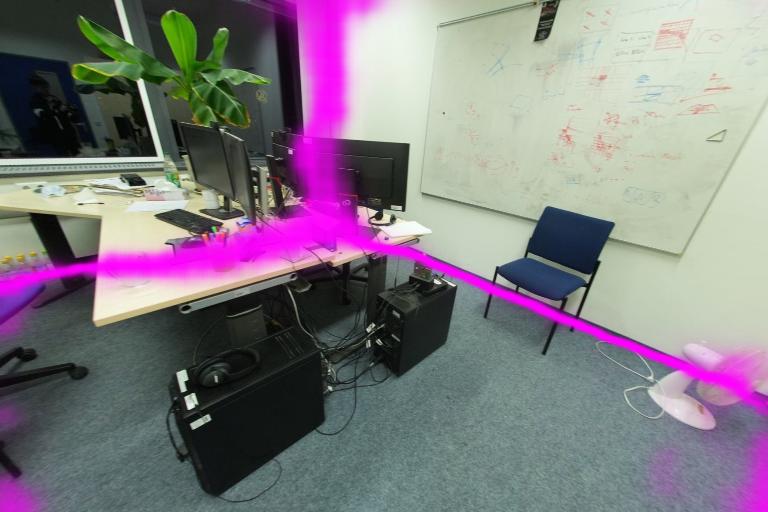}};
        \node at (11.9,-12.2) {\includegraphics[scale=0.12]{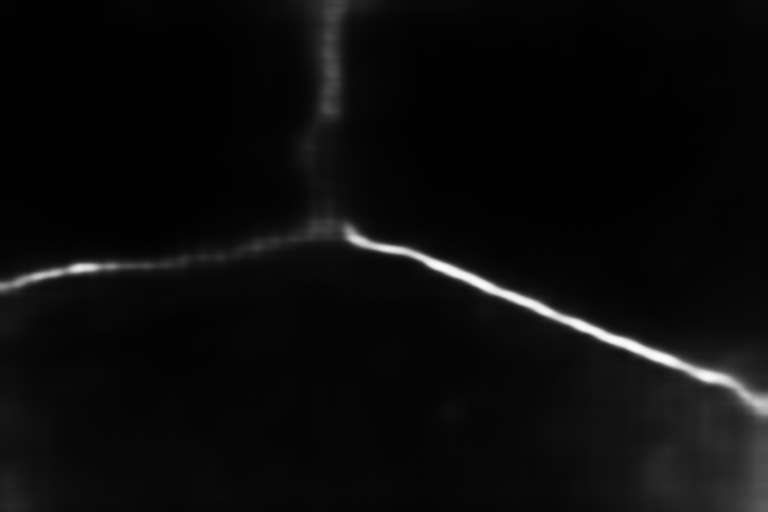}};
    
    \end{tikzpicture}
    \caption{\textbf{Qualitative comparison of predicted feature, edge and confidence maps.} We compare the output from the ResNet-based CNN in Hanning \etal \cite{hanning2025pixcuboid} with our proposed network using a \backbone{} encoder. Results are from the finest scale level. Feature maps are mapped to RGB using PCA. Confidence ranges from low/black to high/white.}
    \label{fig:dino-vs-resnet}
\end{figure*}

\subsection{Cuboid Fitting}
\label{subsec:supp-cuboid-fitting}

In \cref{alg:cuboid-fit} we give the algorithm for fitting a cuboid to a point cloud that was used in
\cref{subsec:comparison-to-simple-baselines}.
%Sec. 5.4.
It randomly samples points (without replacement) from the input point cloud and constructs the cuboid plane-by-plane. Checks for coplanarity have been omitted for brevity.
Note that not every sample of 9 3D-points will yield a cuboid where the points lie on the faces (they will however lie on the infinite planes defined by the faces), thus it is necessary to check the constraints (row 25).
We run the algorithm in a RANSAC loop and pick the cuboid with the maximum number of inliers, defined as points that are within 0.2 m of its faces.
The cuboid is then refined by minimizing the distance between its faces and the inlier points with L-BFGS \cite{liu1989limited}.

\renewcommand{\algorithmicrequire}{\textbf{Input:}}
\renewcommand{\algorithmicensure}{\textbf{Output:}}

\begin{algorithm}
\caption{Fit cuboid to points.}
\label{alg:fit-cuboid-to-points}
\begin{algorithmic}[1]
\Require Point cloud $\bm{P} = \{ \bm{X}_i \}_{i=1}^n$.
\Ensure Cuboid $\mathcal{C}$ or $\emptyset$.
\If{$\left| \bm{P} \right| < 9$}
\State \Return $\emptyset$
\EndIf
\State Sample four points $\bm{X}_i, \bm{X}_j, \bm{X}_k, \bm{X}_l$ from $\bm{P}$.
\State Fit plane $\bm{n}_1^T \bm{X} = a$ to $\bm{X}_i, \bm{X}_j, \bm{X}_k$.
\State Compute offset $b$ such that $\bm{n}_1^T \bm{X}_l = b$.
\State $a, b \gets \min(a, b), \max(a, b)$
%\If{$b < a$}
%\State Swap $a$ and $b$.
%\EndIf
\State $\bm{P} \gets \{ \bm{X_i} \in \bm{P} \mid a < \bm{n}_1^T \bm{X}_i < b \}$
\If{$\left| \bm{P} \right| < 3$}
\State \Return $\emptyset$
\EndIf
\State Sample three points $\bm{X}_m, \bm{X}_n, \bm{X}_o$ from $\bm{P}$.
\State Fit plane $\bm{n}_2^T \bm{X} = c$, orthogonal to $\bm{n}_1$, to $\bm{X}_m, \bm{X}_n$.
\State Compute offset $d$ such that $\bm{n}_2^T \bm{X}_o = d$.
\State $c, d \gets \min(c, d), \max(c, d)$
%\If{$d < c$}
%\State Swap $c$ and $d$.
%\EndIf
%\If{$\exists \bm{X} \in \{ \bm{X}_i, \bm{X}_j, \bm{X}_k, \bm{X}_l \}$ s.t. $\bm{n}_2^T \bm{X} < c$ or $\bm{n}_2^T \bm{X} > d$}
%\State \Return $\emptyset$
%\EndIf
\State $\bm{P} \gets \{ \bm{X_i} \in \bm{P} \mid c < \bm{n}_2^T \bm{X}_i < d \}$
\If{$\left| \bm{P} \right| < 2$}
\State \Return $\emptyset$
\EndIf
\State $\bm{n}_3 \gets \bm{n}_1 \times \bm{n}_2$
\State Sample two points $\bm{X}_p, \bm{X}_q$ from $\bm{P}$.
\State Compute offsets $e, f$ such that $\bm{n}_3^T \bm{X}_p = e, \bm{n}_3^T \bm{X}_q = f$.
\State $e, f \gets \min(e, f), \max(e, f)$
%\If{$f < e$}
%\State Swap $e$ and $f$.
%\EndIf
%\If{$\exists \bm{X} \in \{ \bm{X}_i, \bm{X}_j, \bm{X}_k, \bm{X}_l, \bm{X}_m, \bm{X}_n, \bm{X}_o \}$ s.t. $\bm{n}_2^T \bm{X} < c$ or $\bm{n}_2^T \bm{X} > d$ or $\bm{n}_3^T \bm{X} < e$ or $\bm{n}_3^T \bm{X} > f$}
\For{$\bm{X} \in \{ \bm{X}_i, \dots, \bm{X}_o \}$}
    \If{\textbf{not} $c \leq \bm{n}_2^T \bm{X} \leq d$ \textbf{or not} $e \leq \bm{n}_3^T \bm{X} \leq f$}
        \State \Return $\emptyset$
    \EndIf
\EndFor
%\If{$\bm{n}_2^T \bm{X} < c$ or $\bm{n}_2^T \bm{X} > d$ or $\bm{n}_3^T \bm{X} < e$ or $\bm{n}_3^T \bm{X} > f$ for any sampled point $\bm{X}$}
%\State \Return $\emptyset$
%\EndIf
%\State Construct cuboid $\mathcal{C}$ from the six axis-aligned planes.
\State \Return Cuboid $\mathcal{C}$, built from the six axis-aligned planes.
\end{algorithmic}
\label{alg:cuboid-fit}
\end{algorithm}

\subsection{Impact of Network Backbone}

In this work we propose a \backbone{}-based network as an improvement over the ResNet-101 CNN used in PixCuboid \cite{hanning2025pixcuboid}. To isolate the impact of the new architecture we run both methods with the two different networks and present the results in \cref{tab:dino-resnet}. Here, it is clear that the performance gains of \methodName{} over PixCuboid is not only due to the new network.
On ASE and ScanNet++, \methodName{} with ResNet even outperforms PixCuboid with \backbone{}. PixCuboid achieves the best results on 2D-3D-Semantics. This is not surprising given that the method is specialized on the type of rooms found in this dataset (cuboids), while \methodName{} can handle more general layouts (Manhattan).

\begin{table*}[ht]
\centering
\caption{\textbf{Impact of network backbone.} We run PixCuboid \cite{hanning2025pixcuboid} and \methodName{} (our method) with their ResNet-based network and ours, which is built on \backbone{}.}
\label{tab:dino-resnet}
\begin{tabular}{c l cc cc cc c}
    \toprule
    && \multicolumn{2}{c}{3D} & \multicolumn{2}{c}{Recall} & \multicolumn{2}{c}{Pixel-wise} \\
    \cmidrule(lr){3-4} \cmidrule(lr){5-6} \cmidrule(lr){7-8}
    & Method & {\scriptsize IoU $\uparrow$} & {\scriptsize Chamfer $\downarrow$} & {\scriptsize Wall $\uparrow$} & {\scriptsize Room $\uparrow$} & {\scriptsize Depth $\downarrow$} & {\scriptsize Normal $\uparrow$} \\
    \midrule
    \multirow{4}{*}{\rotatebox{90}{\scriptsize ASE \cite{avetisyan2024scenescript}}} & PixCuboid \scriptsize{(ResNet)} & 68.1 & 0.93 m & 54.9 & 45.5 & 0.57 m & 90.0 \\
    & \methodName{} \scriptsize{(ResNet)} & 82.7 & 0.43 m & 66.8 & 43.2 & 0.29 m & 93.9 \\
    & PixCuboid \scriptsize{(\backbone{})} & 75.9 & 0.68 m & 65.4 & 58.3 & 0.43 m & 92.8 \\
    & \methodName{} \scriptsize{(\backbone{})} ~~~ & \textbf{94.3} & \textbf{0.12 m} & \textbf{89.0} & \textbf{79.0} & \textbf{0.09 m} & \textbf{98.2} \\
    \midrule
    \multirow{4}{*}{\rotatebox{90}{\scriptsize ScanNet \kern-0.6em \cite{yeshwanth2023scannet++}}} & PixCuboid \scriptsize{(ResNet)} & 78.8 & 0.36 m & 58.9 & 26.1 & 0.26 m & 87.8 \\
    & \methodName{} \scriptsize{(ResNet)} & 84.7 & 0.26 m & 66.2 & 31.7 & 0.18 m & 90.7 \\
    & PixCuboid \scriptsize{(\backbone{})} & 81.2 & 0.36 m & 60.8 & 26.7 & 0.25 m & 87.9 \\
    & \methodName{} \scriptsize{(\backbone{})} & \textbf{87.4} & \textbf{0.20 m} & \textbf{69.3} & \textbf{36.3} & \textbf{0.16 m} & \textbf{91.6} \\
    \midrule
    \multirow{4}{*}{\rotatebox{90}{\scriptsize 2D-3D-S \cite{armeni2017joint}}} & PixCuboid \scriptsize{(ResNet)} & 89.0 & 0.18 m & 93.8 & 85.0 & 0.10 m & 96.1 \\
    & \methodName{} \scriptsize{(ResNet)} & 85.4 & 0.28 m & 88.1 & 68.1 & 0.14 m & 95.2 \\
    & PixCuboid \scriptsize{(\backbone{})} & \textbf{92.1} & \textbf{0.11 m} & \textbf{95.8} & \textbf{85.6} & \textbf{0.08 m} & \textbf{96.8} \\
    & \methodName{} \scriptsize{(\backbone{})} & 90.0 & 0.15 m & 92.2 & 80.6 & 0.10 m & 96.4 \\
    \bottomrule
\end{tabular}
\end{table*}

\clearpage
}
{
\maketitle
\begin{abstract}
%Room layout estimation from multiple perspective or panoramic images is a task recently tackled by several methods. However, they either fail to generalize to new domains or make restrictive assumptions about the room shape. Additionally we argue that the repeated structure of buildings has not been fully exploited in prior works. Rooms commonly share dominant directions and floor or ceiling height, in which case they can be estimated together. In this work we propose a novel room layout estimation method called \methodName{}, modeling multi-room layouts as Manhattan polygons that are refined jointly using featuremetric alignment. We furthermore present two new datasets for multi-view room layout estimation and demonstrate the robustness and strong performance of \methodName{}.
%\viktor{Flow-wise this could be sharpened a bit. Perhaps we should also downplay the featuremetric part, maybe (learned optimization instead)}

Estimating room layouts from multi-view imagery is a core task for indoor scene understanding. Existing methods are typically limited either by poor generalization to new datasets or restrictive geometric assumptions of the room shape or camera configuration. Most also estimate rooms independently, failing to exploit shared building structure such as dominant directions, ground plane or ceiling height.

We propose \methodName{}, a multi-room layout estimation method that parameterizes room layouts as Manhattan 3D polygons and optimizes them jointly across multiple rooms. The optimization objective is predicted by a neural network on top of robust pre-trained visual features and trained end-to-end with supervision only on output room layouts. At the same time, camera projection and polygon updates remain explicit and model-based. This separation between learned scoring and geometry improves generalization to new datasets and camera parameters. During optimization, \methodName{} adaptively refines the polygon topology through iterative wall split and merge operations while jointly utilizing structural cues across rooms. We introduce two new multi-view multi-room layout benchmarks by providing layout annotations to existing datasets, and experiments show that \methodName{} %consistently
outperforms prior approaches, both in terms of accuracy and robustness.

Project page: \url{https://ghanning.github.io/\methodName}

\end{abstract}
\section{Introduction}
\label{sec:intro}

Estimating the 3D layout of indoor environments is a core problem in computer vision, with applications in robotics, augmented reality and holistic scene understanding. 
In particular, the goal of \emph{room layout estimation} is to predict the location of the walls, floor and ceiling. 
Many existing methods aim to estimate the layout from a single image, which makes reconstruction difficult and ambiguous.
Recently, Plane-DUSt3R \cite{huang2025unposedsparseviewsroom} and PixCuboid \cite{hanning2025pixcuboid} leverage multiple perspective views for the task.
Plane-DUSt3R retrains the foundation model DUSt3R \cite{wang2024dust3r} to predict the structural planes of the scene, but the method generalizes poorly.
%does not generalize to new datasets.
PixCuboid is an optimization-based approach to
%room
layout estimation centered around featuremetric alignment.
However, it is restricted to single cuboid-shaped rooms which significantly limits the practicality of the method.

In this paper we present a novel multi-view room layout estimation method, dubbed \methodName{}, which shows strong performance across multiple datasets and uses a more general Manhattan world \cite{coughlan1999manhattan} assumption.
Like PixCuboid, our method takes a set of posed perspective images as input
%, where the camera poses resolve the global scale,
and progressively refines an initial room layout by minimizing a learned optimization objective.
Instead of cuboids, we represent layouts as 3D polygons where planes meet at right angles, which makes our method applicable to a larger class of rooms.
For scenes consisting of multiple rooms, we further propose to optimize these together, sharing the same orientation and optionally the floor and ceiling height.
By sharing parameters in the optimization, our method benefits from enhanced convergence and higher accuracy.
%We suggest a network architecture with a \backbone{} \cite{oquab2023dinov2} encoder and train it end-to-end through the optimization.
We also suggest a new network architecture compared to PixCuboid, with a \backbone{} \cite{oquab2023dinov2} encoder, and train it end-to-end through the optimization.
\methodName{} is evaluated on both synthetic and real data and we validate the design in a number of ablation experiments.
The main contributions of this work are:

\begin{figure}[t]
    \centering
    \includegraphics[width=\linewidth]{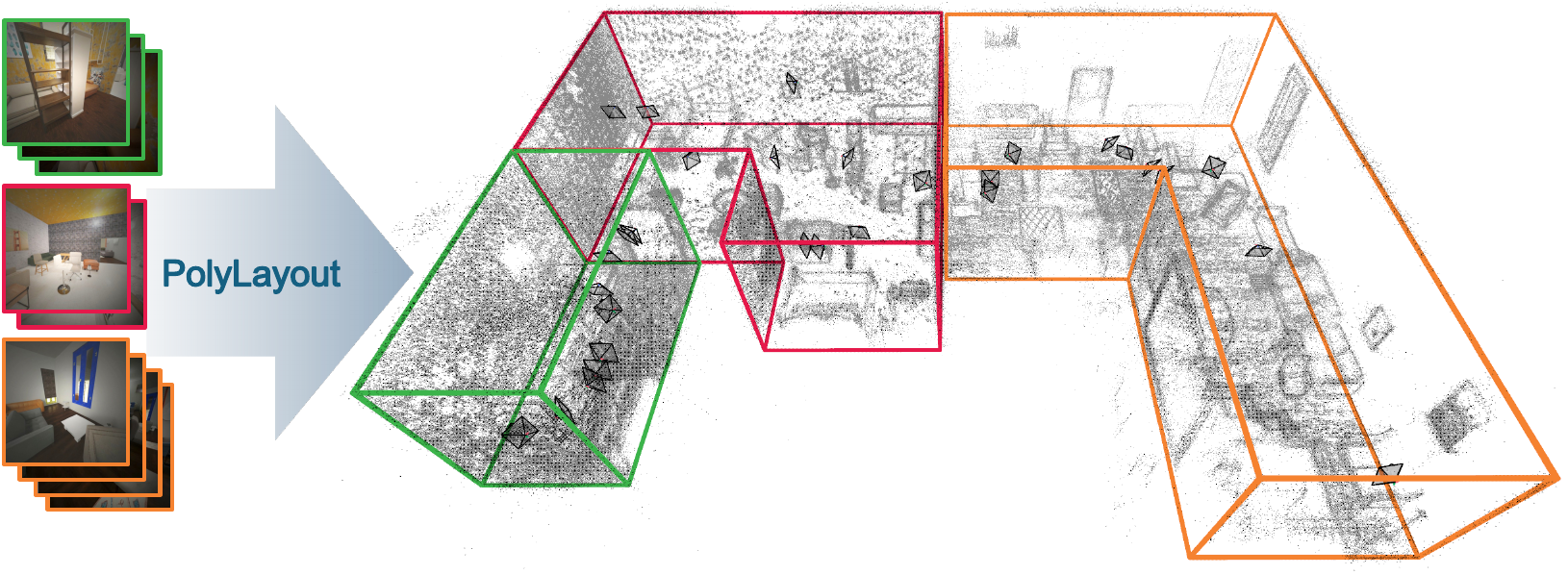}
    \caption{\textbf{Our room layout estimation method \methodName{}} predicts the layout of multiple rooms simultaneously from posed images. By sharing the orientation and floor/ceiling height it achieves higher accuracy than comparable single room methods.}
    \label{fig:teaser}
\end{figure}

\begin{itemize}
    \item We present a multi-view room layout estimation method with strong generalization capabilities, where rooms are represented by Manhattan polygons.
    %\item We extend the cuboid room layout estimation method PixCuboid to more general Manhattan polygon layouts.
    \item We propose to jointly optimize the orientation and floor/ceiling height in multi-room scenes, improving the accuracy of the predicted layouts.
    \item We introduce an adaptive procedure for updating the layout topology during optimization and quantify its effect in ablations.
    %\item We design heuristics for initializing the room layout, as well as mechanisms for adaptively determining the room complexity
    %(number of walls)
    %during optimization.
    \item We create two new datasets, based on Aria Synthetic Environments \cite{avetisyan2024scenescript} and ScanNet++ v2 \cite{yeshwanth2023scannet++} respectively, to benchmark multi-view multi-room layout estimation and show that \methodName{} outperforms competing methods.
    %and show that \methodName{} achieves strong results.
\end{itemize}

\section{Related Work}
\label{sec:related}

%\viktor{Think about how to structure this. Perhaps we can do a split with regression vs. optimization. Where we fall into the optimziation category with (PixCuboid, BADGR, and something else?) }

\textbf{Image-based room layout estimation} is the task of reconstructing the floor, ceiling and walls given one or more views of the indoor environment. Many methods \cite{hedau2009recovering,lee2009geometric,schwing2012efficient,stekovic2020general,nie2020total3dunderstanding,zhang2021holistic} have been proposed to estimate the layout from a single perspective image. However, the unknown scale and limited field-of-view make this a very challenging problem. Other works \cite{zou2018layoutnet,sun2019horizonnet,pintore2021deep3dlayout,wang2021led2} utilize a 360\degree~panorama which gives a more complete coverage of the surroundings but still suffer from scale ambiguity.

Layout estimation from multiple views is a less studied area of research. Flint \etal \cite{flint2011manhattan} reconstruct indoor scenes from video sequences by combining geometric and photometric cues. PSMNet \cite{wang2022psmnet} and GPR-Net \cite{su2023gpr} use a pair of panoramas to predict the room layout with transformer-based networks. MVLayoutNet \cite{hu2022mvlayoutnet} and Pintore \etal \cite{pintore20183d} take as input multiple panoramic views and can reconstruct multi-room scenes. %, but do not make their code publicly available. 
Recently, two methods have been suggested for estimating room layouts from multiple perspective images. Huang \etal \cite{huang2025unposedsparseviewsroom} present Plane-DUSt3R which finds the wall, ceiling and floor planes of an indoor scene from unposed views by leveraging the foundation model DUSt3R \cite{wang2024dust3r}. 
PixCuboid \cite{hanning2025pixcuboid} is an optimization-based method that also uses multiple perspective images but assumes known camera poses and is restricted to cuboid-shaped rooms. It is the most similar method to ours.
%We extend PixCuboid to handle
In contrast to PixCuboid we support more complex room layouts where the number of walls is determined automatically at inference time. Our method can also share parameters between rooms for increased accuracy.
%PixCuboid \cite{hanning2025pixcuboid} is an optimization-based method that also uses multiple perspective images and is the work most similar to ours.
%However, their approach is restricted to single cuboid-shaped rooms.
%In contrast, we consider a more general setting, handling both more complex room layouts (polygons) and estimating the layout for multiple rooms jointly, sharing parameters such as orientation and ceiling height.
%\viktor{We should be clear about novelty w.r.t. pixcuboid, maybe something to the effect of:
%Compared to PixCuboid, our main technical changes are (i) more general layout parameterization, (ii) adaptive topology updates via split/simplify during optimization, and (iii) joint multi-room parameter sharing.}

% Multi-room layout estimation: Pintore et al. \cite{pintore2019automatic} (similar to \citet{pintore20183d}?)
% Floor plan methods, e.g. SALVe \cite{lambert2022salve}

\textbf{Layout estimation from point clouds}. Monte Carlo Scene Search \cite{hampali2021monte} and SceneCAD \cite{avetisyan2020scenecad} are examples of methods that reconstruct the layout of a scene from RGB-D scans.
SceneScript \cite{avetisyan2024scenescript} takes a point cloud obtained from a visual-inertial SLAM system as input and models the layout as a sequence of structured language commands.
The closely related SpatialLM \cite{SpatialLM} fine-tunes an open-source LLM for the same task.
Our method \methodName{} does not require point clouds, instead fitting 3D layout polygons to the images directly.

\textbf{Floor plan reconstruction} is a closely related problem which MonteFloor \cite{stekovic2021montefloor} and RoomFormer \cite{yue2023connecting} solve by extracting 2D polygons from a density map created from a point cloud. BADGR \cite{li2025badgr} reconstructs the floor plan and simultaneously refines the camera poses of input panorama images using diffusion. Our method is richer as it outputs a set of 3D polygons - but they can easily be converted into a 2D floor plan.

\textbf{Room layout assumptions}. Most room layout estimation methods make prior assumptions about the room geometry. Several works \cite{hedau2009recovering,nie2020total3dunderstanding,zhang2021holistic,hanning2025pixcuboid} assume cuboid rooms to simplify the estimation task but are thus unable to reconstruct many real-world building environments. The Manhattan world \cite{coughlan1999manhattan}, where planes are aligned with three principal axes, is a more general model employed by numerous layout estimation methods \cite{zou2018layoutnet,sun2019horizonnet,wang2021led2}. By adding a single-floor, single-ceiling restriction we get the indoor world model \cite{lee2009geometric}. The Atlanta world \cite{schindler2004atlanta} differs from the Manhattan world in that the walls are vertical but not necessarily orthogonal and is used for example by AtlantaNet \cite{pintore2020atlantanet}.
In this paper we utilize the Manhattan world model: room layouts are represented as polygons having edges aligned with the local $x$ and $y$ axes. The Manhattan frame is estimated during optimization. Additionally, our flexible approach allows for sharing the floor and ceiling height in multi-room scenes, resulting in the indoor world model.

%In this paper we utilize the indoor world model: room layouts are represented as polygons having edges aligned with the local $x$ and $y$ axes and share floor and ceiling height. \remi{Shouldn't we say instead that our method is flexible and can use the indoor world model, but can also not make assumptions about the floor/ceiling if we know in advance that the building has a non standard shape?}

% From AtlantaNet paper: "the environment is expected to have horizontal floor and ceiling and vertical walls, but without the restriction of walls meeting at right angles or having a limited number of corners (supporting, e.g., curved walls)."

\textbf{Featuremetric alignment}. Direct alignment of deep features has been used in a number of areas such as point cloud registration \cite{huang2020feature}, camera localization \cite{sarlin2021back} and
%structure from motion
SfM \cite{lindenberger2021pixel}. When trained end-to-end, the learned features offer improved robustness and accuracy. More recently, Hanning \etal \cite{hanning2025pixcuboid} incorporate a featuremetric cost in the refinement of cuboid room layouts. We take inspiration from their work but utilize a vision transformer \cite{dosovitskiy2020vit} (ViT) model for better features.

\section{Method}
\label{sec:method}

\begin{figure}[t]
    \centering    
    \begin{overpic}[width=\linewidth]{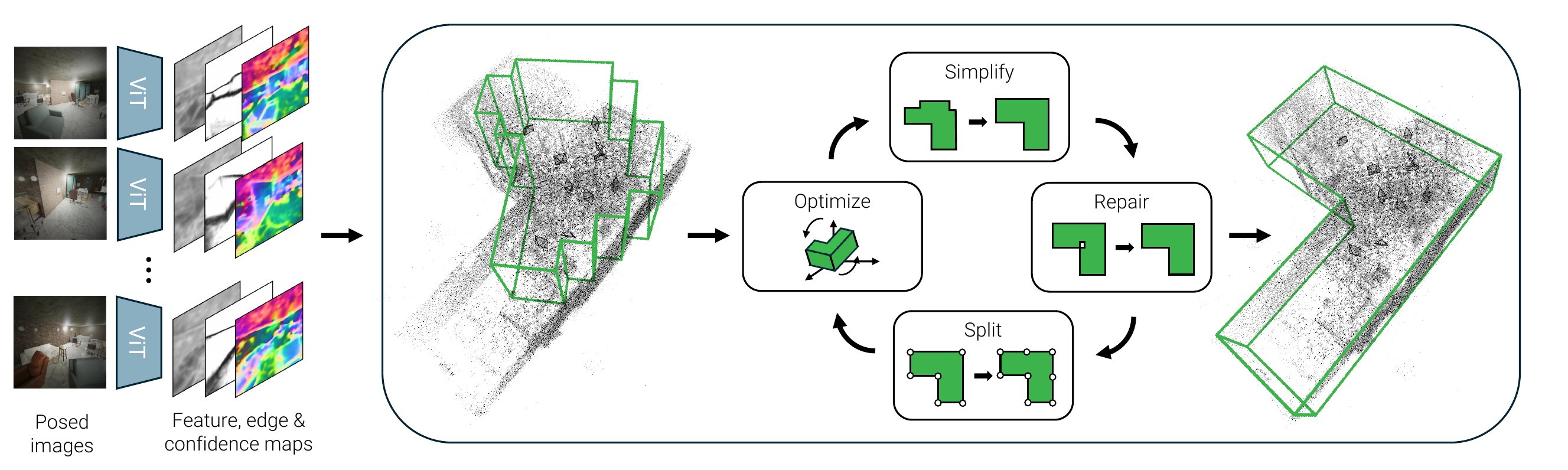}
    \put(38.0,8.0){\scriptsize $\mathcal{P}_{{init}}$}
    \put(90.0,8.0){\scriptsize $\mathcal{P}_{{opt}}$}
    \end{overpic}
    \caption{\textbf{Room layout optimization with \methodName{}.}
    Given posed images, our room layout estimation method \methodName{} refines an initial room layout polygon $\mathcal{P}_{init}$ through successive Levenberg-Marquardt optimization steps, resulting in the final layout $\mathcal{P}_{opt}$. First, a neural network extracts deep features, confidence and edge maps for each input image. The layout is initialized from the known camera poses, and is then optimized by iteratively minimizing the cost function $E(\mathcal{P})$ (\cref{subsec:optimization-of-room-layouts}). After each LM step the polygon is simplified and any self-intersections removed, followed by splitting of the polygon edges (walls). The point cloud is included only for visualization purposes and is not an input to our method.
    }
    \label{fig:overview}
\end{figure}

Our method takes as input a set of images
$\{ \II_i \}_{i=1}^n$
%$\II_1, \II_2, \dots, \II_n$
with known poses $(\bm{R}_i, \bm{t}_i)$ and intrinsics $\bm{K}_i$.
The output is a Manhattan polygon $\mathcal{P}$ where walls
%and ceilings
meet at right angles (\cref{subsec:room-layout-parameterization}), representing the room geometry.
%Our approach jointly optimizes the room layout while determining the complexity of the overall room shape (number of walls).
It is initialized from the camera poses (\cref{subsec:room-layout-initialization}) and then refined through a coarse-to-fine optimization (\cref{subsec:optimization-of-room-layouts}).
During optimization walls can dynamically be removed by simplifying the polygon or added by splitting existing walls (\cref{subsec:updating-the-polygon-layout}).
The method naturally extends to the multi-room setting where parameters are shared across rooms to improve accuracy (\cref{subsec:multi-room-optimization}). We use a network that consists of a ViT encoder and two convolutional decoders
%We use a network design inspired by MoGe-2 \cite{wang2025moge2}
to predict a dense feature map $\FF\in\mathbb{R}^{W\times H\times D}$, an edge map $\EE\in\mathbb{R}^{W\times H}$, as well as two confidence maps $\CF, \CE \in \mathbb{R}^{W\times H}$ for each image $\II\in\mathbb{R}^{W\times H\times 3}$ (\cref{subsec:learning-to-optimize-room-layouts}). An overview 
%of our method
is displayed in \cref{fig:overview}.

\subsection{Room Layout Parameterization}
\label{subsec:room-layout-parameterization}

The room layout is parameterized by the orientation $\bm{R} \in SO(3)$ and a plane offset vector $\bm{d} = \begin{bmatrix} d_1 & d_2 & \cdots & d_p \end{bmatrix} \in \mathbb{R}^p$, where $p \ge 6$. The matrix $\bm{R}$ is the global-to-local rotation and $\bm{d}$ defines the floor, ceiling and wall planes. In the local frame, the floor and ceiling are given by $z = d_1$ and $z = d_2$, respectively. The remaining plane offsets specify the walls of the room in order, alternating between $x$ and $y$: the first wall is thus $x = d_3$, the second $y = d_4$, and so on. We show an example of this representation in \cref{fig:polygon} (left). There is an even number of wall planes, at least four. In the case of four planes, the shape is simply a cuboid.

\subsection{Optimization of Room Layouts}
\label{subsec:optimization-of-room-layouts}

Our method is based on optimizing an initial layout through successive Levenberg-Marquardt \cite{levenberg1944method,marquardt1963algorithm} steps in a coarse-to-fine manner, with the cost function
\begin{equation}
    E(\mathcal{P}) = 
    E_{feat}(\mathcal{P}) + \alpha E_{edge}(\mathcal{P}) + \beta E_{VP}(\mathcal{P}) + \gamma E_{per}(\mathcal{P}).
\end{equation}
We adapt the three cost functions used in \cite{hanning2025pixcuboid} to the polygon case: $E_{feat}$ which measures multi-view consistency of warped deep image features, $E_{edge}$ that aligns the layout edges with predicted edge maps, and $E_{VP}$ which compares the three vanishing points defined by the layout to detected line segments. More specifically, the featuremetric cost
\begin{equation}
    {E}_{feat}(\mathcal{P}) =
    \sum_{i,j} \sum_k w_{ijk} \rho \left( \left\| \FF_i[\xx_{ik}] - \FF_j\left[\mathcal{W}_{i\to j}(\xx_{ik}, \mathcal{P})\right] \right\|^2 \right),
\end{equation}
where $\{ \xx_{ik} \}$ are image points sampled in image $\II_i \in \mathbb{R}^{W \times H \times 3}$ and $\mathcal{W}_{i\to j}$ denotes the warping to image $\II_j$ via the polygon $\mathcal{P}$. $\FF_i, \FF_j \in \mathbb{R}^{W \times H \times D}$ are dense feature maps extracted by a neural network and $[\cdot]$ represents lookup with sub-pixel interpolation. $\rho$ is a robust loss function and
\begin{equation}
    w_{ijk} = \CF_i[\xx_{ik}] \CF_j[\mathcal{W}_{i\to j}(\xx_{ik}, \mathcal{P})]    
\end{equation}
are per-point weights interpolated from confidence maps $\CF_i, \CF_j \in \mathbb{R}^{W \times H}$. The points $\{ \xx_{ik} \}$ are sampled from the probability map $\CF_i^\kappa$, $\kappa \in \mathbb{R}$, without replacement.
In contrast to the cuboid case it is possible to have self-occlusions with polygons, which need to be taken care of when warping between images.

The edge cost $E_{edge}$ is computed by sampling 3D points $\{ \bm{X}_k \}$ along the edges of the polygon and projecting into each view:
\begin{equation}
    E_{edge}(\mathcal{P}) =
    \sum_i \sum_k w_{ik} \EE_i[\Pi_i(\bm{R}_i \bm{X}_k + \bm{t}_i)]^2.
\end{equation}
Here, $\Pi_i: \mathbb{R}^3 \to \mathbb{R}^2$ denotes the projection using the known intrinsics $\bm{K}_i$ and $\EE_i \in \mathbb{R}^{W \times H}$ is the predicted edge map. As with the featuremetric cost we use a confidence image $\CE_i$ from which the weights $w_{ik}$ are interpolated.
%
%\begin{equation}
%    w_{ik} = \CE_i[\Pi_i(\bm{R}_i \bm{X}_k + \bm{t}_i)].
%\end{equation}

Thanks to our Manhattan assumption the polygon $\mathcal{P}$ has only three vanishing points (VPs), given for each view by the columns of $\bm{R}_i \bm{R}^T = \begin{bmatrix} \bm{v_{i,1}} & \bm{v_{i,2}} & \bm{v_{i,3}} \end{bmatrix}$.
The vanishing point cost $E_{VP}$ measures the consistency \cite{tardif2009non} of these VPs and line segments $\{ \bm{l}^i_{k} \}$ detected in the images $\II_i$:
%For each image we extract line segments with \cite{pautrat2023deeplsd} and use a vanishing point cost $E_{VP}$ that measures the consistency \cite{tardif2009non} of these VPs and the line segments $\{ \bm{l}^i_{k} \}$ found in the images $\II_i$:
%
\begin{equation}
    E_{VP}(\mathcal{P}) =
    \sum_i \sum_k \min \left( \min_{m\in\{1,2,3\}} D_{VP}(\bm{l}^i_{k}, \bm{v}_{i,m}), \tau \right)^2.
\end{equation}
Line segments are softly assigned to only one VP by taking the minimum distance
\begin{equation}
    D_{VP}(\bm{l}, \bm{v}) = | \hat{\bm{l}}^T \bm{l}_1 | / \sqrt{\hat{l}_1^2 + \hat{l}_2^2}
\end{equation}
between one of the segment end points $\bm{l}_1$ and the line $\hat{\bm{l}} = \bar{\bm{l}} \times \bm{v} = \begin{bmatrix} \hat{l}_1 & \hat{l}_2 & \hat{l}_3 \end{bmatrix}$ passing through its midpoint $\bar{\bm{l}}$ and the vanishing point $\bm{v}$. The distance is capped at $\tau$ to handle line segments that do not align with any VP.

In this work we additionally introduce a perimeter cost $E_{per}$ to penalize complex polygon layouts $\mathcal{P}$:
\begin{equation}
    E_{per}(\mathcal{P}) = |d_3 - d_{p-1}| + |d_4 - d_p| + \sum_{i=4}^{p-1} |d_{i+1} - d_{i-1}|. 
\end{equation}
In cases where only a subset of walls are visible in the images, this cost induces a slight shrinking bias that prevents the unobserved regions from diverging.

\textbf{Coarse-to-fine optimization}: Our network outputs feature, confidence and edge maps on three different scale levels, with gradually increasing resolution (1/16, 1/4 and 1/1 of the input image size). We start the optimization at the coarsest level and initialize subsequent scales with the output from the previous. % one.

\subsection{Learning to Optimize Room Layouts}
\label{subsec:learning-to-optimize-room-layouts}

The network is trained end-to-end, supervised on the optimized room layouts at each scale. We use a set of ground truth 2D-3D correspondences $\{ (\bm{x}_{ik}^{GT}, \bm{X}_{ik}^{GT}) \}$ with points on the walls, floor and ceiling to compute the loss
\begin{equation}
    \mathcal{L}(\mathcal{P}) =
    \frac{1}{N}
    \sum_{i,j} \sum_k
    \rho \left(
    \| \mathcal{W}_{i\to j}(\bm{x}_{ik}^{GT}, \mathcal{P}) - \Pi_j (\bm{R}_j \bm{X}_{ik}^{GT} + \bm{t}_j) \|^2
    \right)
\end{equation}
and propagate the gradients back through the unrolled optimization process. We apply the loss to the optimized layout at each scale, but only if the optimization succeeded at the previous level ($\mathcal{L}(\mathcal{P})$ below a threshold) to prevent oversmoothing of the fine feature maps. During training, we
%use room layouts with four walls (cuboids), and
disable the VP and perimeter costs ($\beta = \gamma = 0$) as they contains no learned components. The polygons are initialized by randomly rotating and shifting the planes of the ground truth layouts for each room.
As in previous works \cite{sarlin2021back,hanning2025pixcuboid}, we learn per-parameter LM dampening factors along with the weights of the feature extractor network. Walls share a single dampening factor. Similar to \cite{hanning2025pixcuboid}, we also pre-train the edge maps with a weighted MSE loss, where the target images are line renderings of the ground truth polygon edges.
Our network has a \backbone{} ViT encoder and two convolutional heads for the feature and edge maps. The architecture is detailed in the supplementary material.
%\todo{The network architecture is based on MoGe-2, utilizing a pre-trained \backbone{} \cite{oquab2023dinov2} backbone for the encoder and two convolutional heads for the feature and edge maps.}

\subsection{Room Layout Initialization}
\label{subsec:room-layout-initialization}

\begin{figure}[t]
    \centering
    \subfloat{
    \begin{tikzpicture}[scale=1.0]
        \draw[line width=1.0pt, draw=black] 
            (0,0) -- 
            (3,0) -- 
            (3,1) -- 
            (1,1) -- 
            (1,2.5) -- 
            (0,2.5) -- 
            cycle;
        \draw[->, darkgray, thick, >={Latex}] (-0.3,-0.3) -- (0.7,-0.3) node[right] {$x$};
        \draw[->, darkgray, thick, >={Latex}] (-0.3,-0.3) -- (-0.3,0.7) node[above] {$y$};
        \node[font=\footnotesize, rotate=90] at (3.25, 0.50) {$x = d_3$};
        \node[font=\footnotesize] at (2.12, 1.26) {$y = d_4$};
        \node[font=\footnotesize, rotate=90] at (1.25, 1.84) {$x = d_5$};
        \node[font=\footnotesize] at (0.54, 2.76) {$y = d_6$};
        \node[font=\footnotesize, rotate=90] at (0.25, 1.28) {$x = d_7$};
        \node[font=\footnotesize] at (1.52, 0.26) {$y = d_8$};
    \end{tikzpicture}
    \quad \quad \quad
    %\vspace{-0.2cm}
    \includegraphics[width=.4\linewidth]{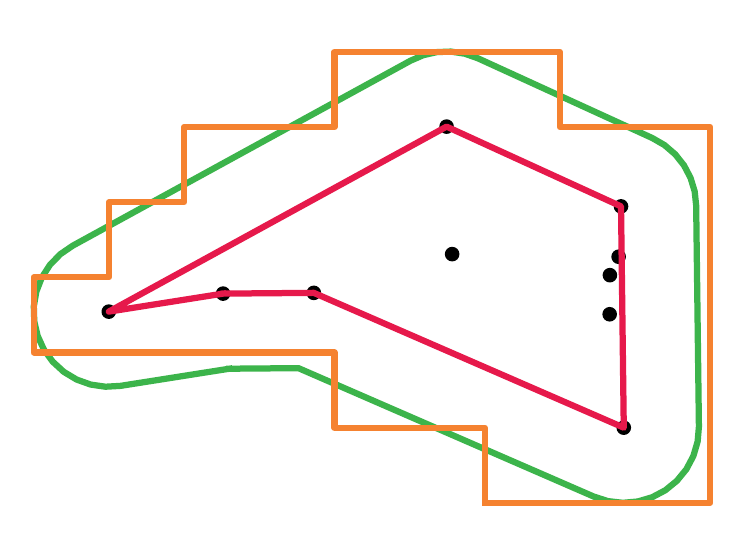}
    %\vspace{-0.7cm}
    }
    \caption{\textbf{Left}: The walls of the room layout polygon $\mathcal{P}$ are defined by axis-aligned planes, alternating between $x$ and $y$. \textbf{Right}: To initialize $\mathcal{P}$ from the positions of the cameras (black points) we first compute the concave hull (\textcolor{RedColor}{\bf red}), then buffer it (\textcolor{GreenColor}{\bf green}) and convert it to a Manhattan polygon through rasterization (\textcolor{OrangeColor}{\bf orange}). %\viktor{draw camera frustum (triangles), ceiling,floor d1,d2}
    }
    \label{fig:polygon}
\end{figure}

During inference we initialize $\mathcal{P}$ from the known camera poses. The mean of the camera $y$ axes gives an up vector ($z$), which together with two randomly sampled basis vectors orthogonal to $z$ form the rotation matrix $\bm{R}$. As in \cite{hanning2025pixcuboid} we then take a few LM optimization steps minimizing $E_{VP}(\mathcal{P})$ to get a more accurate initial orientation. In the local $xy$ plane defined by $\bm{R}$ an $\alpha$-shape \cite{edelsbrunner2003shape} is computed around the camera positions and expanded by a fixed distance $\delta$. The shape is converted into a Manhattan polygon by rasterizing it and then tracing the outline of the resulting binary image. Floor and ceiling heights are set so that the minimum distance to the camera centers is $\delta$. The proposed initialization strategy is visualized in \cref{fig:polygon} (right image). 

% As the number of walls in the room is not known at inference time we initialize $\mathcal{P}$ with a large number of walls, approximating a circle around the cameras (\cref{fig:polygon-initialization}). The mean of the camera $y$ axes gives a up vector ($z$), which together with two randomly sampled basis vectors orthogonal to $z$ form the rotation matrix $\bm{R}$. In the local $xy$ plane defined by $\bm{R}$ we compute the centroid of the camera centers and set the radius to be maximum distance between this centroid and the cameras. Points are then sampled equiangularly on the circle defined by the centroid and the given radius, and walls added to connect adjacent points. Finally the floor and ceiling height is set to fit all the cameras with a certain margin, resulting in the initial room layout polygon. During the optimization process walls are subsequently removed due to simplification or added by splitting as described in \cref{subsec:updating-the-polygon-layout}.

\subsection{Updating the Polygon Layout}
\label{subsec:updating-the-polygon-layout}

After each optimization step at inference time, we first check if any camera center is outside the room layout, in which case the closest walls (or floor/ceiling) are moved outwards as needed. We then simplify the layout polygon using a variant of the Visvalingam-Whyatt algorithm \cite{visvalingam1993line}. In the original formulation the importance of a polygon vertex is given by the area spanned by itself and its two neighbors, and the vertex with minimum importance is iteratively removed. To maintain the Manhattan property of our polygon we adjust the algorithm to remove two adjacent walls at a time, with importance equal to the area of the rectangle spanned by the two walls. Only walls that have converged (small $|\Delta d_i|$) are eligible for removal. Any self-intersections that might have been introduced by the LM step or simplification are removed using the Shapely \cite{Gillies_Shapely_2025} library. After optimization on the coarse and medium scale levels, walls longer than a threshold are iteratively split, up to a specified maximum number of planes $p_{max}$.

\subsection{Multi-room Optimization}
\label{subsec:multi-room-optimization}

The parameterization described in \cref{subsec:room-layout-parameterization} allows for joint optimization of multiple rooms with shared orientation and floor/ceiling height. 
Alternative parameterizations, \eg via rotation/translation/size $(\bm{R}, \bm{t}, s_x, s_y, s_z)$, entangle the parameters for each plane with the translation, making them harder to share between different rooms.
Sharing parameters across rooms greatly reduces the degrees of freedom and improves convergence, resulting in more accurate layout predictions.
We argue that rooms within a building having the same orientation is extremely common, and show in ablation experiments (\cref{subsec:ablation-experiments}) how accuracy improves when our method exploits this fact. Note that wall locations can also be shared, but it is not explored in this work.
%
%In practice we first run the coarse-to-fine optimization with the same orientation for all rooms. Shared floor, ceiling and wall planes are then identified. We greedily merge planes that are closer than a threshold ($|d_i - d_j| < d_\Delta$), as long as they correspond to the same axis and have the same normal direction. Finally the optimization is rerun on the finest scale. \methodName{} hence makes no single-floor or single-ceiling assumption but instead detects shared planes automatically at runtime.
%
We further assume prior knowledge of the image assignment for multi-room scenes, \ie we know which images belong to which room.

\section{Experimental Setup}
\label{sec:experimental-setup}

We use the following parameters when evaluating \methodName{}: $\alpha = 0.05$, $\beta = 40$ and $\gamma = 5 \times 10^{-4}$ (on the finest scale, $\gamma = 0$). 256 points are sampled in each view to compute the featuremetric cost $E_{feat}$ and $\rho$ is the generalized loss function from \cite{barron2019general}. For the edge cost $E_{edge}$ we sample 40 points uniformly along every polygon edge. The VP distance threshold $\tau$ is set to 0.05 and we detect line segments with DeepLSD \cite{pautrat2023deeplsd}, using up to 100 segments per image for $E_{VP}$. We initialize the room layout polygon as described in \cref{subsec:room-layout-initialization}, and run five iterations of VP optimization to align it with detected line segments. The number of LM steps per scale level is 15. After each optimization step the polygon is simplified using an importance threshold of 0.5.
Unless otherwise stated the orientation $\bm{R}$ and floor/ceiling height ($d_1, d_2$) are shared between the rooms in multi-room scenes.
% The orientation is shared between the rooms in multi-room scenes and we identify common plane offsets as outlined in \cref{subsec:multi-room-optimization}.
The setup is more extensively described in our supplementary material.
%For more details
% about the experimental setup, 
%see our supplementary material.
% (\cref{subsec:supp-implementation-details}).

\subsection{Training}
\label{subsec:training}

To train \methodName{}'s neural network we follow the procedure given in \cite{hanning2025pixcuboid}, but swap their ResNet-101 \cite{he2016deep} based CNN for a \backbone{} backbone (ViT-S/14) and two convolutional heads.
We take their training set, 391 cuboid rooms from ScanNet++ v2, and extend it with 107 manually annotated Manhattan room layouts from the same dataset.
%We train on the same data: 391 cuboid rooms from ScanNet++ v2.
More details and a comparison between the two models can be found in the supplementary material.
%(\cref{fig:feat-edge-conf-maps}).
%We noticed that the pre-training converged much faster with \backbone{}, so the number of epochs is decreased from 10 to 3.
%The edge maps are first pre-trained for 3 epochs.
%In the second stage of training we use a batch size of 2 and set the initial learning rate to $3.33 \times 10^{-7}$ for the encoder and $3.33 \times 10^{-6}$ for the convolutional heads. After 10 epochs it is reduced by a factor of 5 and training continues for another 5 epochs. The two training stages take around 4 h and 72 h, respectively, on one NVIDIA TITAN V GPU.
%with 12 GB of memory.

\subsection{Datasets}
\label{subsec:datasets}

\begin{figure}[t]
    \centering
    \begin{tikzpicture}
    \begin{axis}[
        width=0.65\columnwidth,
        height=0.3\columnwidth,
        xlabel={Number of rooms},
        ylabel={Number of scenes}, % normalized
        ybar, % Sets the chart type to a bar chart
        bar width=10pt, % Adjust the width of the individual bars
        enlargelimits=0.15, % Adds a bit of padding around the bars and axes
        enlarge y limits=false,
        ymin=0.0,
        bar width=8pt,
        yticklabels=\empty,
        ytick=\empty,
        axis line style={->},
        legend style={
            at={(1.35,0.25)},
            anchor=north,
        },
        axis lines*=left,
        symbolic x coords={1, 2, 3, 4, 5},
        xtick=data, % Place ticks exactly at the data points
        legend cell align={left},
        legend image code/.code={
            \draw [#1] (0cm,-0.09cm) rectangle (0.11cm,0.18cm); },
        ]
    ]
    
    % --- Data Series 1: ASE validation st ---
    \addplot[
        fill=RedColor,
    ] coordinates {
        (1, .40) % (Number of Rooms, Number of Scenes)
        (2, .32)
        (3, .17)
        (4, .07)
        (5, .04)
    };
    \addlegendentry{ASE validation}
    
    % --- Data Series 2: ASE test set ---
    \addplot[
        fill=GreenColor,
    ] coordinates {
        (1, .43)
        (2, .25)
        (3, .22)
        (4, .09)
        (5, .01)
    };
    \addlegendentry{ASE test}

    % --- Data Series 3: ScanNet++ test set ---
    \addplot[
        fill=OrangeColor,
    ] coordinates {
        (1, 0.85)
        (2, 0.075)
        (3, 0.05)
        (4, 0.0125)
        (5, 0.0125)
    };
    \addlegendentry{ScanNet++ test}
    
    \end{axis}

    \draw (6.09, 0.70) rectangle (10.39, 2.30) {};
    %\draw[fill=white] (2.1, 3.5) rectangle (5.9, 3.7) {};
    \node at (8.15,2.05) {\small Percentage of cuboid rooms};

    % ------------------------------------------------------------------
    % --- PGF-PIE CHART CODE FOR INSET PLOTS ---
    
    \begin{scope}[shift={(current axis.center)}, xshift=102pt, yshift=8pt]
        %\draw[gray,step=0.25] (-0.5,-.5) grid (.5, .5);
        \clip (-.6,-.6) rectangle (.6, .6);
        \pie[
            radius=0.04\columnwidth,
            color={RedColor, RedColor!40},
            hide number,
            hide label,
        ]{63.05/Cuboid, 36.95/Non-cuboid}
        %\node[font=\scriptsize\bfseries, text=white] at (-0.02, 0.18) {Cuboid};
        \node[font=\footnotesize\bfseries, text=white] at (-0.04, 0.22) {63};
        \node[font=\footnotesize\bfseries, text=white] at (0.13, -0.22) {37};
    \end{scope}

    \begin{scope}[shift={(current axis.center)}, xshift=142pt, yshift=8pt]
        \clip (-.6,-.6) rectangle (.6, .6);
        \pie[
            radius=0.04\columnwidth,
            color={GreenColor, GreenColor!40},
            hide number,
            hide label,
        ]{60.0/Cuboid, 40.0/Non-cuboid}
        %\node[font=\scriptsize\bfseries, text=white] at (-0.02, 0.18) {Cuboid};
        \node[font=\footnotesize\bfseries, text=white] at (-0.04, 0.22) {60};
        \node[font=\footnotesize\bfseries, text=white] at (0.12, -0.22) {40};
    \end{scope}

    \begin{scope}[shift={(current axis.center)}, xshift=182pt, yshift=8pt]
        \clip (-.6,-.6) rectangle (.6, .6);
        \pie[
            radius=0.04\columnwidth,
            color={OrangeColor, OrangeColor!40},
            hide number,
            hide label,
        ]{27.885/Cuboid, 72.115/Non-cuboid}
        %\node[font=\scriptsize\bfseries, text=white] at (0.22, 0.18) {Cub};
        \node[font=\footnotesize\bfseries, text=white] at (0.19, 0.21) {19};
        \node[font=\footnotesize\bfseries, text=white] at (-0.05, -0.22) {81};
    \end{scope}
    % ------------------------------------------------------------------
    
    \end{tikzpicture}
    \caption{\textbf{Distribution of the number of rooms} in our ASE and ScanNet++ datasets. The pie charts show the ratio of cuboid (dark color) versus non-cuboid rooms.}
    \label{fig:dataset-distribution}
\end{figure}

%\subsubsection{Aria Synthetic Environments}

\textbf{Aria Synthetic Environments} \cite{avetisyan2024scenescript} (ASE) is a synthetic dataset containing 100,000 multi-room scenes with simulated camera trajectories and ground truth 3D floor plans. It has a mix of cuboid-shaped rooms and more general Manhattan layouts.
%Scenes have a single floor/ceiling height.
All rooms within a scene have the same floor and ceiling height.
We create new validation and test sets for multi-view room layout estimation by randomly selecting 100 scenes for each set. 
Our method is tuned on the validation set. 
From the ground truth, which is given as a list of walls, doors and windows, we derive the geometry of every room by connecting adjacent walls.
Five sets of images, consisting of 10 images per room, are sampled for all scenes. We make sure that the images have sufficient coverage of the scene by employing a visibility-based sampling. 
Details are given in the supplementary material.

% (\cref{subsec:supp-image-sampling}).

%\subsubsection{ScanNet++}

\textbf{ScanNet++}: For evaluation on real data we manually annotate the room layouts of 80 scenes from ScanNet++ v2 \cite{yeshwanth2023scannet++}. Both single- and multi-room scenes are included, with cuboid-shaped rooms as well as more general layouts that are not necessarily Manhattan or have shared ceiling height. The annotation is performed by fitting planes to mesh vertices selected by the user in a 3D graphical user interface. When planes have been fitted to the floor, ceiling and walls the annotation program automatically finds the intersections between the planes to create the ground truth room layout. 
%Due to limitations of our program 
For the annotation we only consider rooms that have a single horizontal ceiling and flat walls.
A small number of additional rooms are excluded due to incomplete lidar scans. Overall, only a minor portion of the dataset is discarded.
%We only consider rooms with a single, horizontal ceiling and flat walls. Some rooms are also skipped because the semantic mesh is incomplete.
%\viktor{Can we say anything about how many rooms this is? Small minority?}
%
For each scene three sets of images are sampled, with 10 DSLR images per room. The image sampling process is described in our supplementary material.
% (\cref{subsec:supp-image-sampling}).
%
Note that these 80 scenes annotated by us are not part of the ScanNet++ dataset presented in \cite{hanning2025pixcuboid}, and do not overlap with our training data.

% When re-training the neural network (\cref{subsec:training-details}) we only use their training set, consisting of cuboid-shaped rooms.

%\subsubsection{2D-3D-Semantics}

\textbf{2D-3D-Semantics}: We additionally make use of the 2D-3D-Semantics \cite{armeni2017joint} dataset from \cite{hanning2025pixcuboid}, with 160 cuboid rooms. For every room there are two panoramas, split into four perspective images each. To support our multi-room setup the rooms are grouped together within each building.

The distribution of the number of rooms per scene and their type for ASE and ScanNet++ is shown in \cref{fig:dataset-distribution}. We will make the image sets and ground truth
%room
layouts for these two datasets available together with code for evaluation.
%For the following experiments, all hyperparameters are selected on ASE validation set and then kept fixed for all test evaluations.

\subsection{Metrics}
\label{subsec:metrics}

We adopt the 3D (IoU, Chamfer distance), pixel-wise (depth RMSE, normal angle recall at 10\degree) and mean prediction time metrics proposed in \cite{hanning2025pixcuboid}, but compute them per scene instead of per room for ASE and ScanNet++. In addition, we define wall and room recall as follows. For each wall in the ground truth layouts a grid of 3D points is sampled with 0.25 m spacing. The minimum distance between each point and the predicted room layout is then computed. If more than 90\% of the points are within 0.25 m the wall is considered to be successfully captured by the prediction. This "wall recall" metric is averaged over all ground truth walls. The "room recall" is the proportion of rooms with 100\% wall recall.

\section{Results}
\label{sec:results}

\subsection{Room Layout Estimation}
\label{subsec:room-layout-estimation}

\textbf{Baselines on ASE and ScanNet++}: 
We compare \methodName{} with the two recent multi-view room layout estimation methods Plane-DUSt3R \cite{huang2025unposedsparseviewsroom} and PixCuboid \cite{hanning2025pixcuboid}. We further include SceneScript \cite{avetisyan2024scenescript} as a representative point-based approach, as well as the floor plan reconstruction method RoomFormer \cite{yue2023connecting}.
Plane-DUSt3R and PixCuboid are run individually for each room while SceneScript, RoomFormer and \methodName{} are run on a per-scene basis. We use the authors’ official implementations and pre-trained network weights.
%The input images are undistorted beforehand, and for ASE they are rotated 90\degree~to upright orientation. 

%SceneScript has three variants, but only the point cloud version is public, which we use for the experiments. We generate dense COLMAP~\cite{schoenberger2016sfm,schoenberger2016mvs} point clouds with fixed poses and filter outliers by keeping points inside or within 1 m of the ground truth layouts. For comparison, we also use ASE’s semi-dense SLAM point clouds, noting that our test scenes come from ASE training data and were seen during training.

SceneScript has three variants: using a point cloud, posed images, or both. Since only the point cloud version is publicly available, we use that. We generate point clouds via dense COLMAP~\cite{schoenberger2016sfm,schoenberger2016mvs} reconstruction with fixed camera poses. As SceneScript is sensitive to outliers, we filter the cloud to keep only points inside the ground truth room layouts or within 1 m of them. For reference, we also report SceneScript on the semi-dense ASE point clouds produced by SLAM using all scene images. Note also that the method was trained on ASE with these point clouds, and that our test set is a subset of the ASE training scenes.

%SceneScript comes in three different flavors, taking either a point cloud, posed images, or a combination of both as input. However, only the point cloud version has been publicly released, so we use that one. In order to get the points we perform a dense COLMAP \cite{schoenberger2016sfm,schoenberger2016mvs} reconstruction for every set of images, with fixed camera poses. Through our experiments we noticed that SceneScript is sensitive to outliers and we therefore filter the point cloud, keeping the points that are inside the ground truth room layouts or within 1 m. For reference we also run SceneScript on the semi-dense point cloud included in the ASE dataset, generated by a SLAM algorithm using the full set of images in the scene. Note though that our test set is created from the training scenes of ASE, meaning that the model has seen these point clouds during training.

For Plane-DUSt3R the metric-scale version is used. We tried fixing the camera poses but got very poor results, so we instead align the predicted poses to the ground truth using COLMAP's model aligner tool.
%utilize their predicted poses and the ground truth intrinsics to compute the pixel-wise metrics. The Chamfer distance and wall and room recall are calculated after first aligning the predicted poses with the ground truth ones using COLMAP's model aligner tool.

Both SceneScript and Plane-DUSt3R output walls which do not form a closed volume, and so we do not report the IoU for these methods. Additionally, the floor and ceiling are removed from the ground truth layout when computing the Chamfer distance for SceneScript and Plane-DUSt3R.

RoomFormer takes a density map as input, which we create by projecting the point cloud from the dense COLMAP reconstruction onto the ground plane. The output is a set of 2D polygons that are lifted into 3D using the ground truth floor and ceiling height. We use the checkpoint trained on Structured3D \cite{zheng2020structured3d}.

\textbf{Baselines on 2D-3D-Semantics}: 
The availability of panoramic images makes comparison with panorama-based methods Deep3DLayout \cite{pintore2021deep3dlayout}, LED\textsuperscript{2}-Net \cite{wang2021led2} and PSMNet \cite{wang2022psmnet} possible.
We further include results for Total3D-Understanding \cite{nie2020total3dunderstanding} and Implicit3DUnderstanding \cite{zhang2021holistic}, which estimate the layout from a single perspective image.
We do not compare with MVLayoutNet \cite{hu2022mvlayoutnet} or Pintore \etal \cite{pintore20183d} as there is no publicly available code.
For 2D-3D-S, all methods except \methodName{} are run per room. When computing 3D and recall metrics for single-view methods we only consider the prediction with maximum IoU against the ground truth layout. All metrics are calculated per room for this dataset.%, as in \cite{hanning2025pixcuboid}.

Certain methods (SceneScript, Plane-DUSt3R, RoomFormer) may predict empty layouts. In this case we consider the IoU and recall to be zero, but skip computing the Chamfer distance as it is not well-defined. Similarly, we skip views in which the predicted layout is not visible for the pixel-wise metrics. The prediction time metric measures only the inference time and does not include for example the dense COLMAP reconstruction (SceneScript, RoomFormer)
%, density map rendering (RoomFormer)
or DeepLSD line detection (PixCuboid, \methodName{}).

\begin{table*}[t]
\centering
\caption{\textbf{Room layout estimation} on our Aria Synthetic Environments and ScanNet++ v2 test sets and the cuboid-shaped spaces of 2D-3D-Semantics.}
\label{tab:room-layout-estimation}

\def\cubeicon{\raisebox{-0.08cm}{\includegraphics[width=0.32cm]{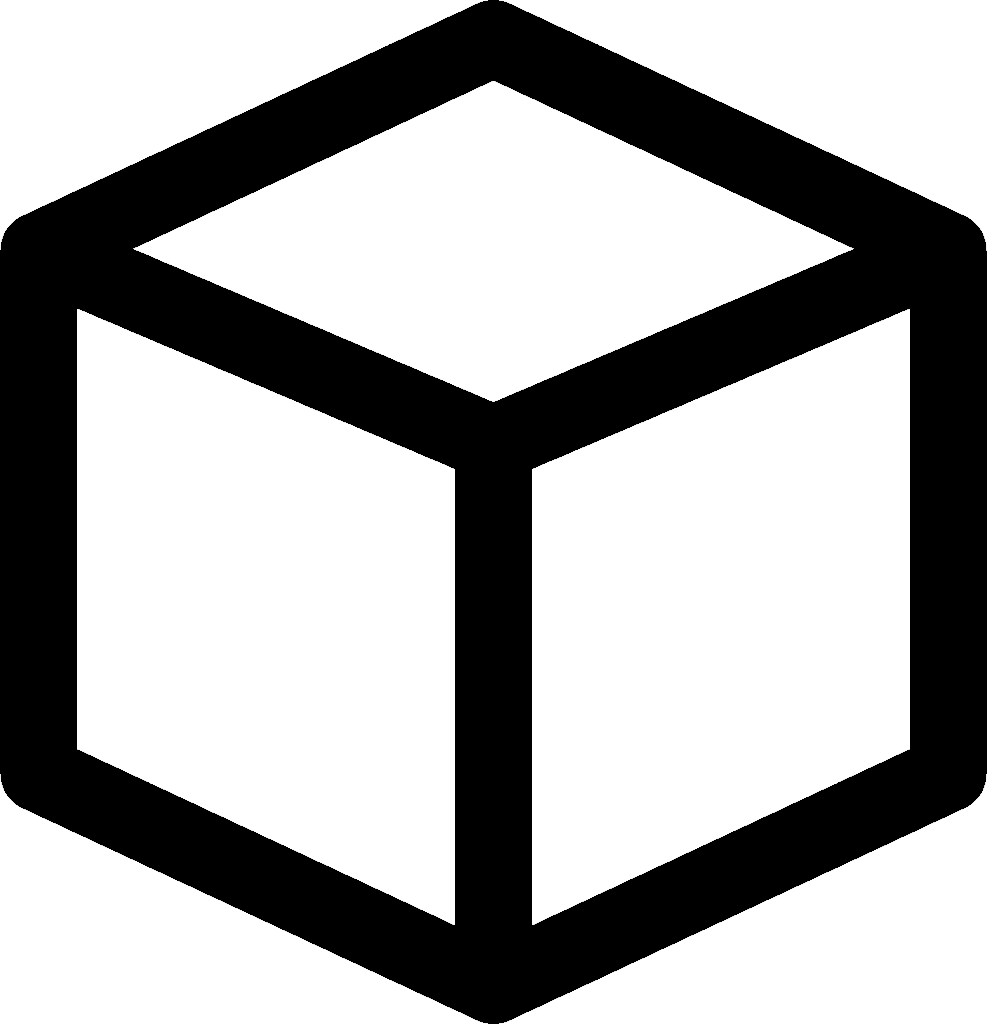}}}

\begin{threeparttable}
\begin{tabular}{c l cc cc cc c}
    \toprule
    && \multicolumn{2}{c}{3D} & \multicolumn{2}{c}{Recall} & \multicolumn{2}{c}{Pixel-wise} \\
    \cmidrule(lr){3-4} \cmidrule(lr){5-6} \cmidrule(lr){7-8}
    & Method & {\scriptsize IoU $\uparrow$} & {\scriptsize Chamfer $\downarrow$} & {\scriptsize Wall $\uparrow$} & {\scriptsize Room $\uparrow$} & {\scriptsize Depth $\downarrow$} & {\scriptsize Normal $\uparrow$} & {\scriptsize Time $\downarrow$} \\
    \midrule
    \multirow{6}{*}{\rotatebox{90}{\scriptsize ASE \cite{avetisyan2024scenescript}}} & SceneScript\tnote{\textdagger}~ \cite{avetisyan2024scenescript} {\tiny ECCV24} & N/A & 0.12 m\tnote{$\ddagger$} & 96.1 & 91.0 & 0.07 m & 98.6 & 5.31 s \\
    \cdashline{2-9}\noalign{\vskip\belowrulesep}
    & SceneScript \cite{avetisyan2024scenescript} {\tiny ECCV24} & N/A & 2.44 m\tnote{$\ddagger$} & 16.7 & 9.1 & 1.18 m & 64.5 & 7.45 s \\
    & Plane-DUSt3R \cite{huang2025unposedsparseviewsroom} {\tiny ICLR25} & N/A & 23.16 m\tnote{$\ddagger$} & 0.1 & 0.0 & 1.90 m & 19.7 & 105.60 s \\
    & PixCuboid \cite{hanning2025pixcuboid} {\tiny ICCVW25} & 68.1 & 0.93 m & 54.9 & 45.5 & 0.57 m & 90.0 & 1.01 s \\
    & RoomFormer \cite{yue2023connecting} {\tiny CVPR23} & 13.8 & 2.37 m & 5.0 & 0.5 & 1.75 m & 67.8 & \textbf{0.06 s} \\
    & \methodName{} (ours) & \textbf{94.3} & \textbf{0.12 m} & \textbf{89.0} & \textbf{79.0} & \textbf{0.09 m} & \textbf{98.2} & 5.48 s \\
    \midrule
    \multirow{5}{*}{\rotatebox{90}{\scriptsize ScanNet \cite{yeshwanth2023scannet++}}} & SceneScript \cite{avetisyan2024scenescript} {\tiny ECCV24} & N/A & 1.80 m\tnote{$\ddagger$} & 11.9 & 0.3 & 0.80 m & 55.8 & 6.16 s \\
    & Plane-DUSt3R \cite{huang2025unposedsparseviewsroom} {\tiny ICLR25} & N/A & 16.82 m\tnote{$\ddagger$} & 0.2 & 0.0 & 1.06 m & 18.2 & 57.72 s \\
    & PixCuboid \cite{hanning2025pixcuboid} {\tiny ICCVW25} & 78.8 & 0.36 m & 58.9 & 26.1 & 0.26 m & 87.8 & 0.87 s \\
    & RoomFormer \cite{yue2023connecting} {\tiny CVPR23} & 10.3 & 1.61 m & 3.4 & 0.0 & 0.98 m & 48.6 & \textbf{0.11 s} \\
    & \methodName{} (ours) & \textbf{87.4} & \textbf{0.20 m} & \textbf{69.3} & \textbf{36.3} & \textbf{0.16 m} & \textbf{91.6} & 3.54 s \\
    \midrule
    \multirow{7}{*}{\rotatebox{90}{\scriptsize 2D-3D-Semantics \cite{armeni2017joint}}} & Total3D \cite{nie2020total3dunderstanding} {\tiny CVPR20} & 31.8 & 1.57 m & 6.9 & 0.0 & 1.41 m & 44.1 & 0.32 s \\
    & Implicit3D \cite{zhang2021holistic} {\tiny CVPR21} & 31.6 & 1.59 m & 5.6 & 0.0 & 1.51 m & 37.6 & \textbf{0.15 s} \\
    & Deep3DLayout \cite{pintore2021deep3dlayout} {\tiny TOG21} & 58.8 & 0.68 m & 16.7 & 0.0 & 0.44 m & 52.6 & 1.30 s \\
    & LED\textsuperscript{2}-Net \cite{wang2021led2} {\tiny CVPR21} & 68.7 & 0.45 m & 20.6 & 5.0 & 0.40 m & 34.6 & 0.71 s \\
    & PSMNet \cite{wang2022psmnet} {\tiny CVPR22} & 43.5 & 1.04 m & 7.3 & 0.0 & 0.63 m & 51.4 & 2.32 s \\
    & PixCuboid \cite{hanning2025pixcuboid} {\tiny ICCVW25} & 89.0 & 0.18 m & \textbf{93.8} & \textbf{85.0} & \textbf{0.10 m} & 96.1 & 0.42 s \\
    & \methodName{} (ours) & \textbf{90.0} & \textbf{0.15 m} & 92.2 & 80.6 & \textbf{0.10 m} & \textbf{96.4} & 1.11 s \\
    %\cdashline{2-9}\noalign{\vskip\belowrulesep}
    %& \methodName{} (ours) \cubeicon &  & & & & & & \\
    \bottomrule
\end{tabular}
\begin{tablenotes}
    \small
    \item [\textdagger] Uses the semi-dense point cloud of the ASE dataset, computed from all images in the scene. SceneScript is trained on ASE, from which our test set is extracted.
    \item [$\ddagger$] Calculated after first removing the floor \& ceiling from the ground truth layout.
    %\item [\resizebox{0.22cm}{!}{\cubeicon}] Restricting the polygon shape to be a cuboid, similar to PixCuboid. Note that all rooms in this dataset are cuboid-shaped.
\end{tablenotes}
\end{threeparttable}
\end{table*}

%\subsubsection{Aria Synthetic Environments}
%\textbf{Aria Synthetic Environments}:
\textbf{Results}: 
We first evaluate on our ASE test set (\cref{tab:room-layout-estimation}, top). 
Using the semi-dense SLAM point cloud, SceneScript can reconstruct the room geometry very accurately (first row). With point clouds created just from the 10 images per room there is a significant drop in performance (second row). Plane-DUSt3R, trained on Structured3D \cite{zheng2020structured3d}, does not generalize and achieves near-zero recall and worse pixel-wise metrics than SceneScript (third row). While only capable of estimating cuboid rooms, PixCuboid outperforms the two previously mentioned methods (fourth row). RoomFormer is by far the fastest but struggles to reconstruct floor plans from the relatively low density point cloud (fifth row). \methodName{} (last row) is considerably more accurate than the competing methods, even rivaling SceneScript with the semi-dense point cloud on some metrics.

%\subsubsection{ScanNet++}
%\textbf{ScanNet++}:
Next, the same methods are applied to the new ScanNet++ v2 test set with manually annotated room layouts (\cref{tab:room-layout-estimation}, center). This dataset contains rooms with more detailed geometry and shorter wall segments, making the recall problem more challenging. Results follow the same trend as for ASE, with PixCuboid and \methodName{} being the top two methods. The difference between them is smaller and can be explained by the fact that many of the rooms are not cuboids, but nearly cuboid-shaped.

%\subsubsection{2D-3D-Semantics}
%\textbf{2D-3D-Semantics}:
Finally, we evaluate on the cuboid-shaped spaces of 2D-3D-Semantics (\cref{tab:room-layout-estimation}, bottom). As the rooms do not have the same floor and ceiling height we run \methodName{} with only shared orientation. Our method scores better on the 3D and pixel-wise metrics than PixCuboid, despite the latter being developed specifically for cuboid room layout estimation. No other method show competitive results on this dataset.
% We also try running \methodName{} with cuboids (last row), setting $\gamma = 0$ and initializing the layouts as in \cite{hanning2025pixcuboid}. With this setup the best performance is achieved, thanks to the shared cuboid orientation and our stronger network.

\begin{figure}[tp]
    \centering
    \begin{tikzpicture}[scale=1.0]
        \node at (0.0,0.0) {\includegraphics[scale=0.08]{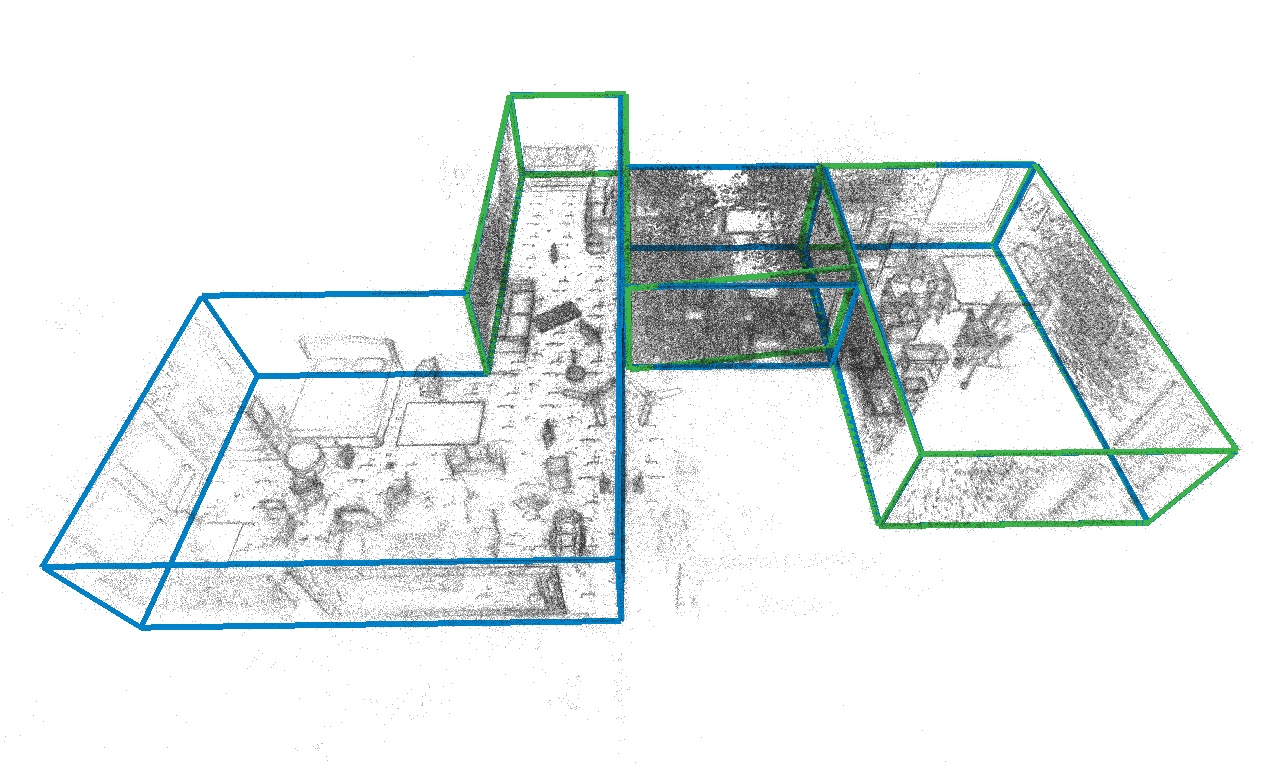}};
        \node[align=center] at (0.0,-1.4) {SceneScript};

        \node at (3.7,0.0) {\includegraphics[scale=0.08]{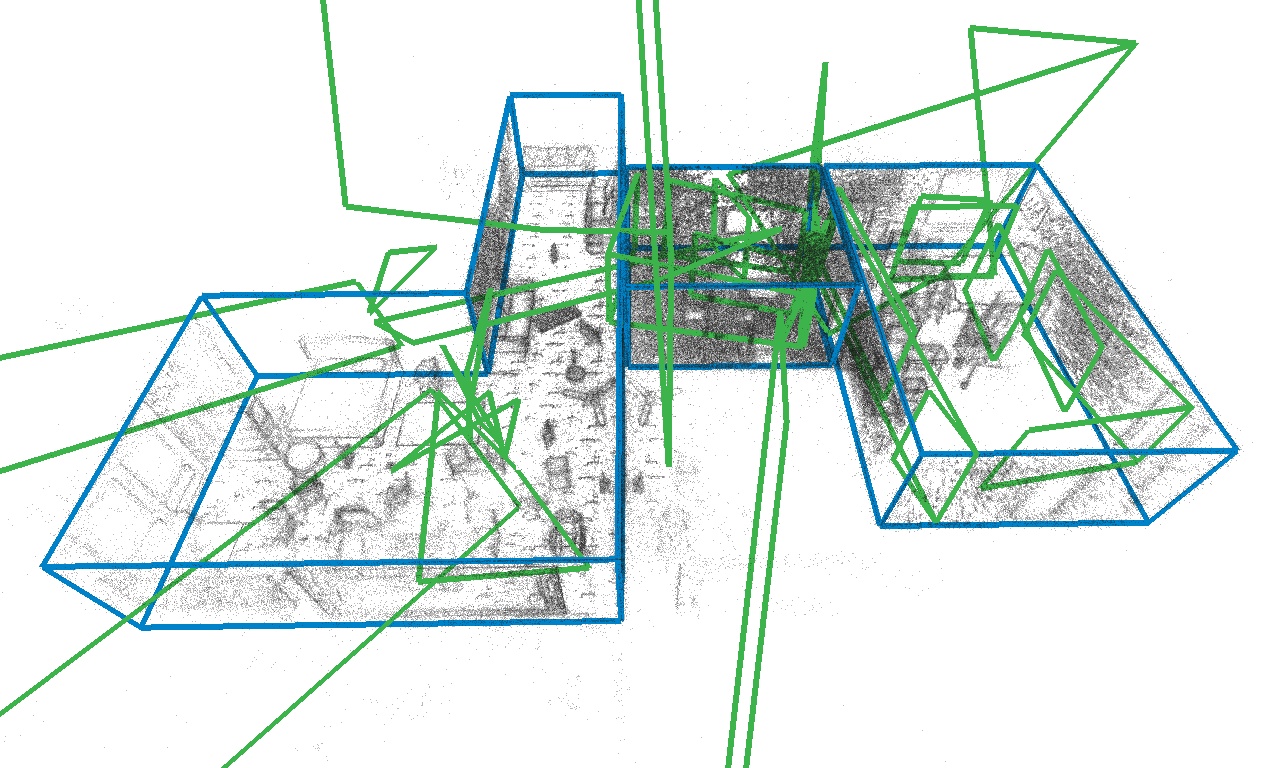}};
        \node[align=center] at (3.7,-1.4) {Plane-DUSt3R};
        
        \node at (0.0,-2.8) {\includegraphics[scale=0.08]{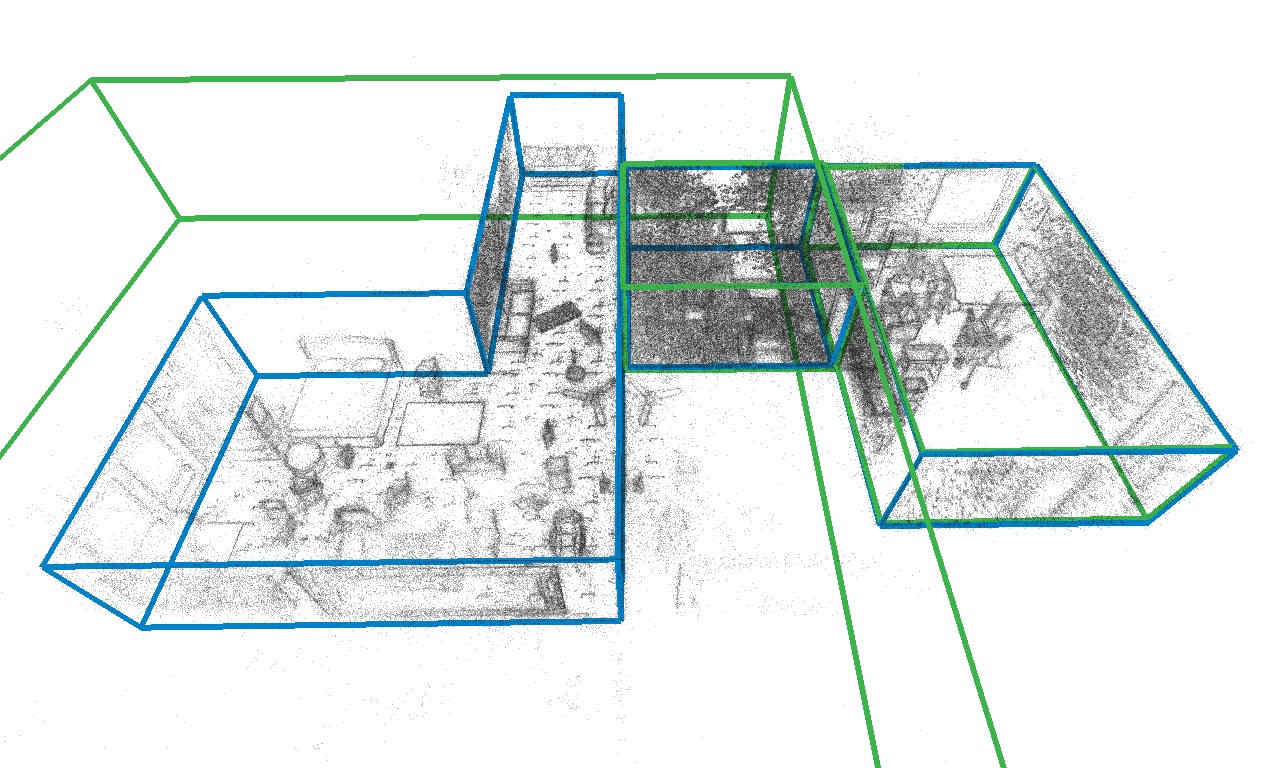}};
        \node[align=center] at (0.0,-4.2) {PixCuboid};

        \node at (3.7,-2.8) {\includegraphics[scale=0.08]{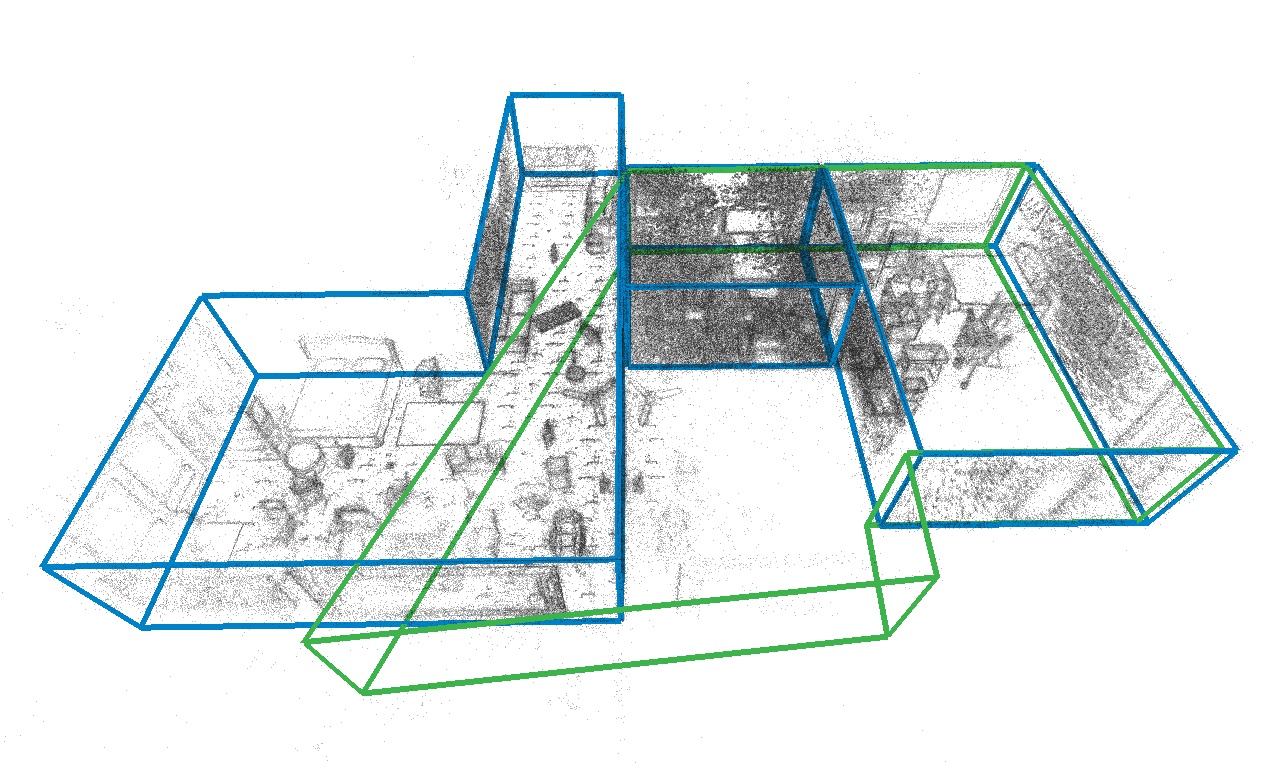}};
        \node[align=center] at (3.7,-4.2) {RoomFormer};

        \node at (7.8,-1.40) {\includegraphics[scale=0.105]{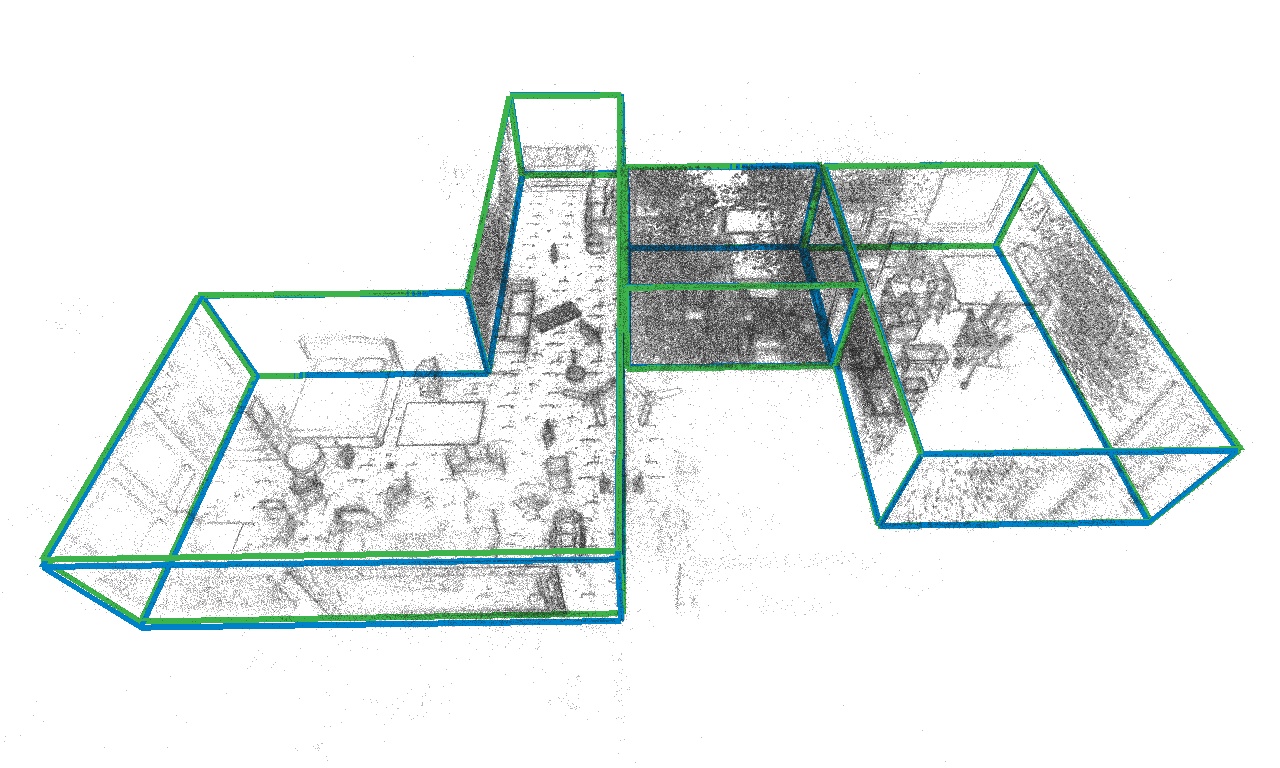}};
        \node[align=center] at (7.8,-2.95) {\methodName{}};

        \node at (0.0,-5.9) {\includegraphics[scale=0.08]{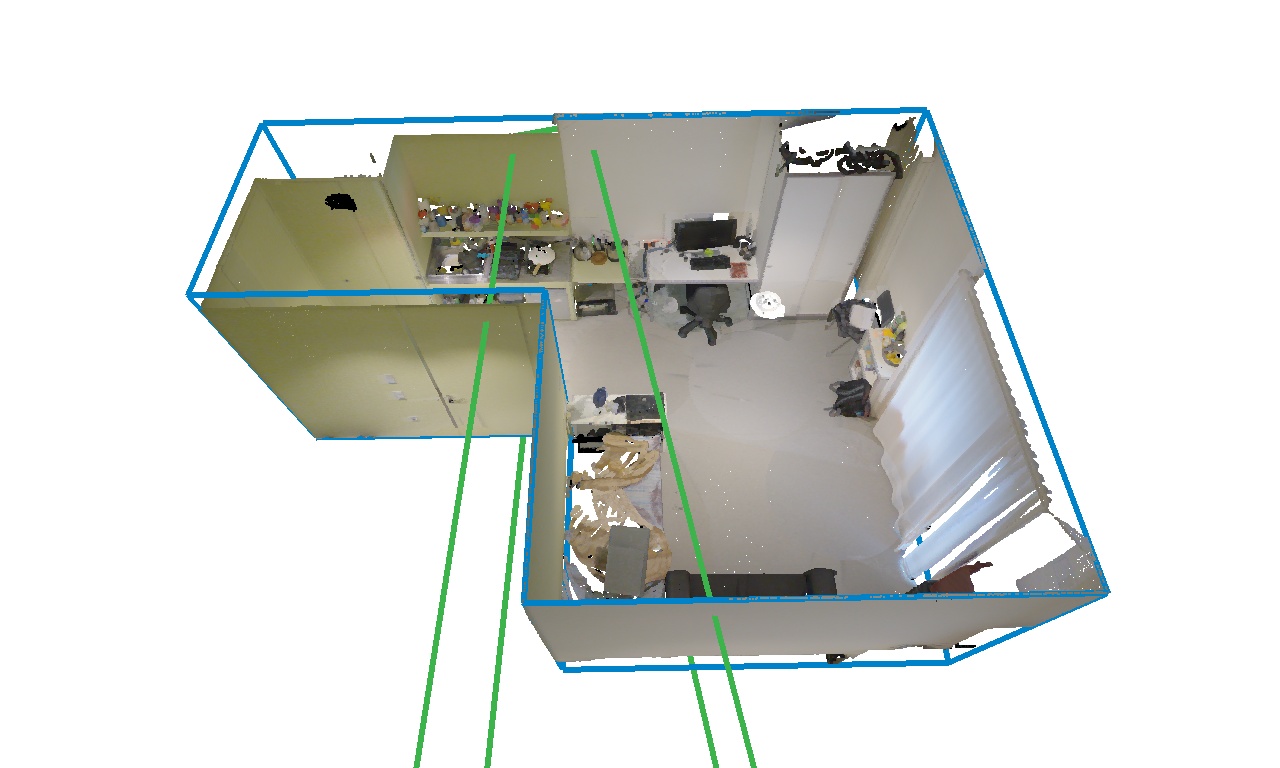}};
        \node[align=center] at (0.0,-7.3) {SceneScript};

        \node at (3.7,-5.9) {\includegraphics[scale=0.08]{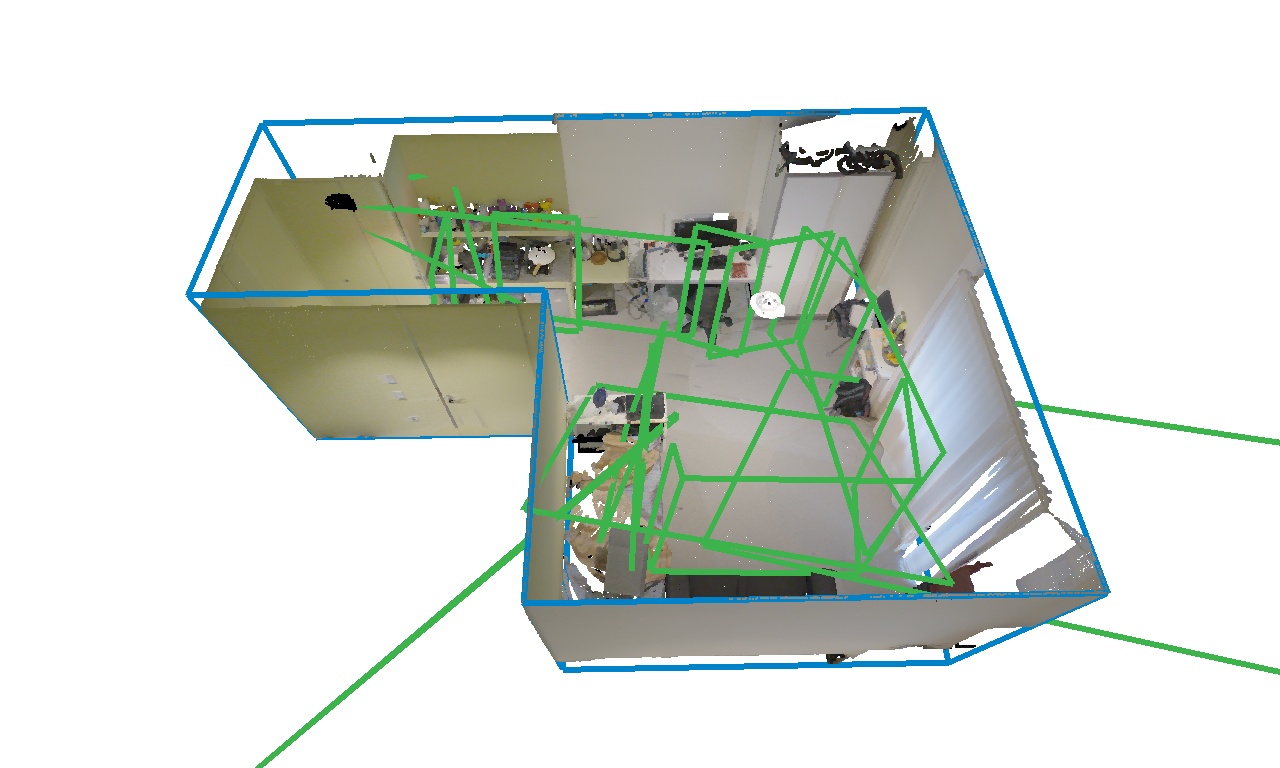}};
        \node[align=center] at (3.7,-7.3) {Plane-DUSt3R};
        
        \node at (0.0,-8.7) {\includegraphics[scale=0.08]{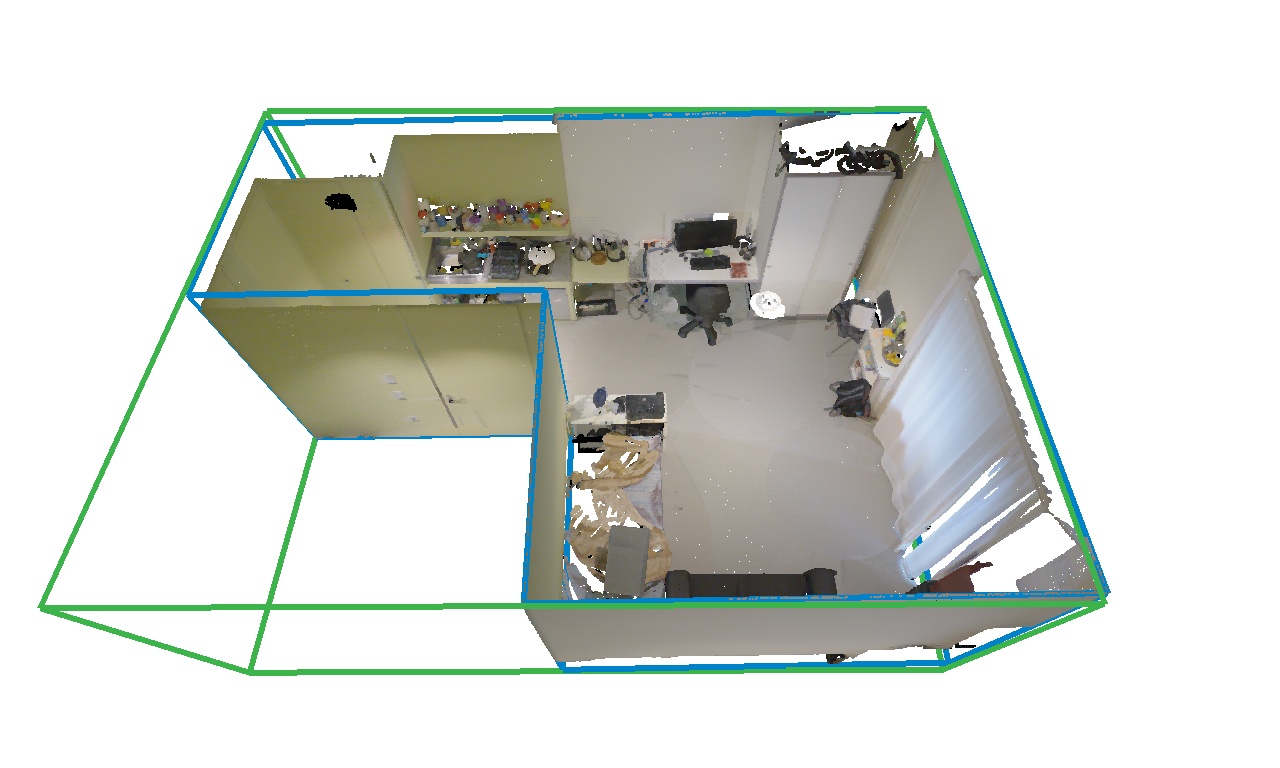}};
        \node[align=center] at (0.0,-10.1) {PixCuboid};

        \node at (3.7,-8.7) {\includegraphics[scale=0.08]{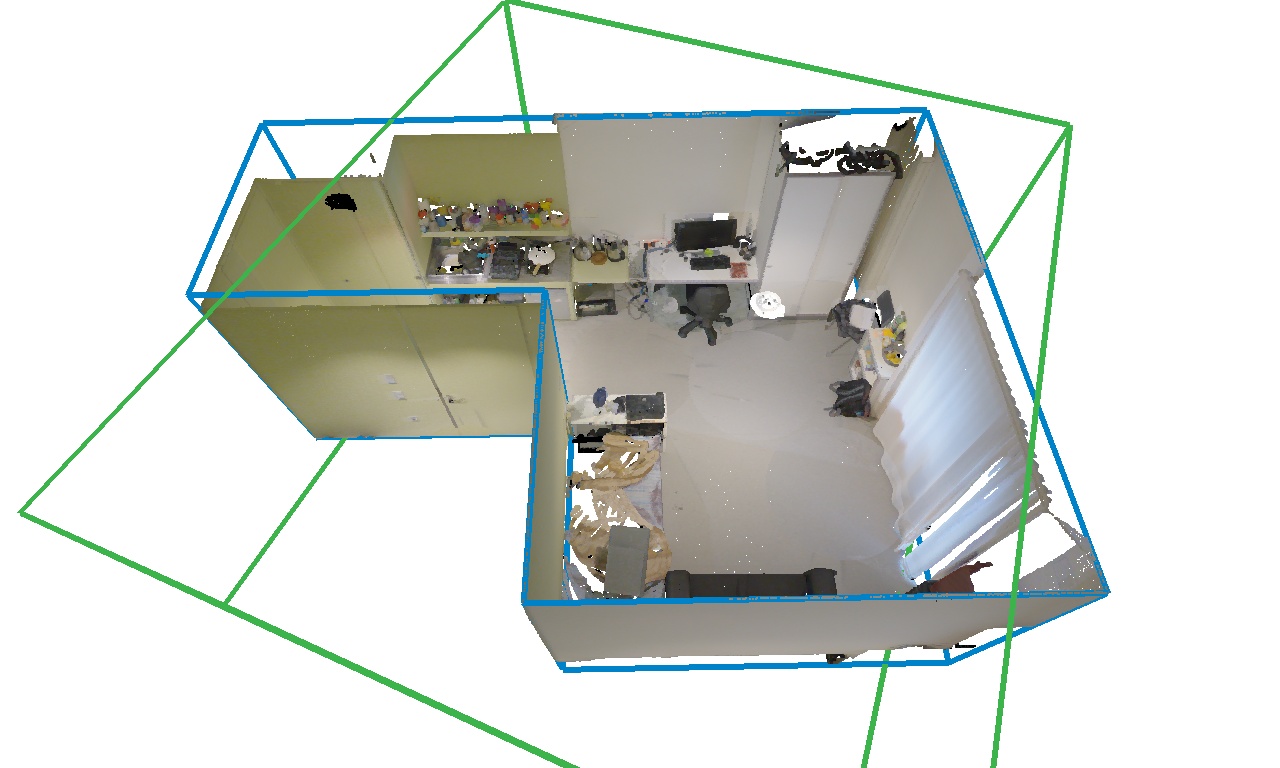}};
        \node[align=center] at (3.7,-10.1) {RoomFormer};

        \node at (7.8,-7.3) {\includegraphics[scale=0.105]{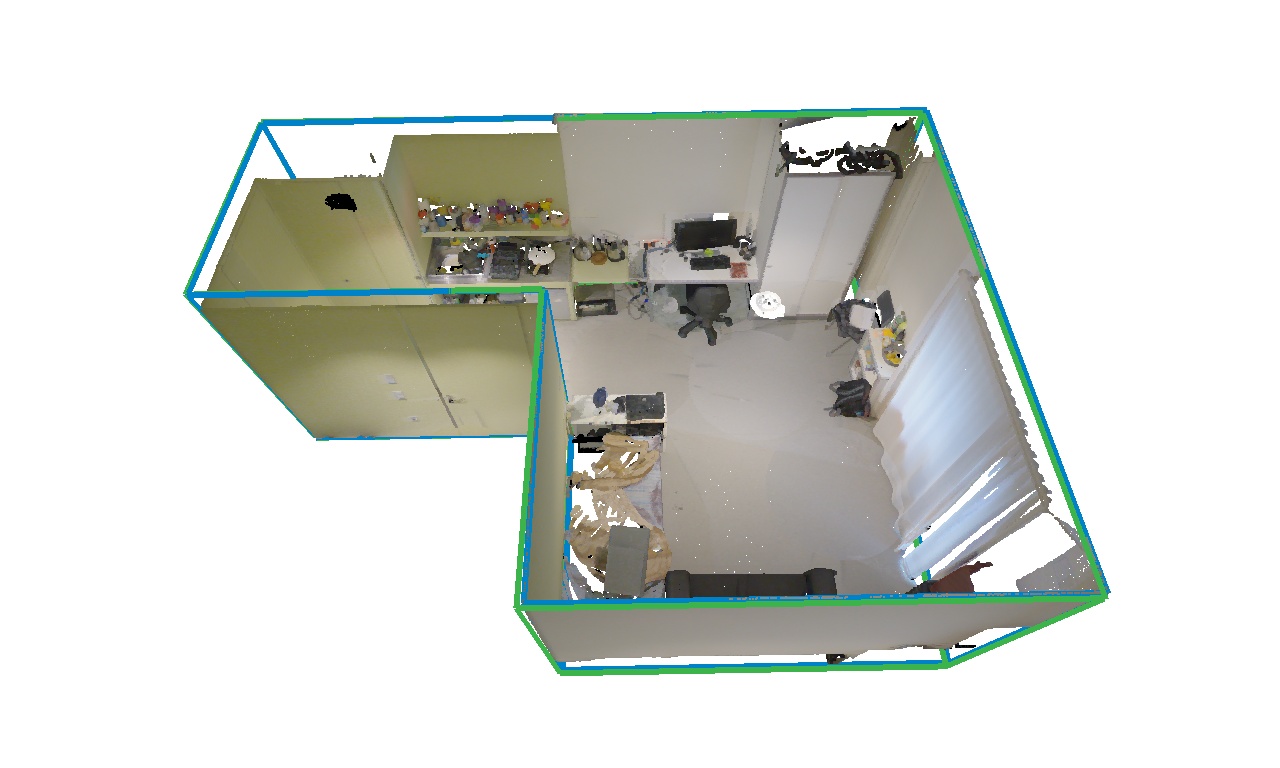}};
        \node[align=center] at (7.8,-8.85) {\methodName{}};

        \draw[dashed,thin] (-1.0,-4.7)--(9.0,-4.7);
    \end{tikzpicture}
    \caption{\textbf{Qualitative comparisons} of predicted room layouts for one scene in each of our ASE and ScanNet++ test sets. Predictions are shown in \textcolor{PredColor}{\bf green} and the ground truth layouts in \textcolor{GTColor}{\bf blue}. The point cloud is displayed only for visualization purposes.
    %More examples can be found in our supplementary material.
    }
    \label{fig:layouts}
\end{figure}

In \cref{fig:layouts} we visualize the predicted room layouts for one scene in each of the ASE and ScanNet++ test sets. Additional examples and failure cases can be found in the supplementary material. 

\subsection{Ablation Experiments}
\label{subsec:ablation-experiments}

We perform several ablation experiments on the ASE test set to validate our design decisions. Parameter sharing across rooms is first considered, comparing single and multi-room layout estimation. In \cref{tab:ablation-experiments} (top section) it can be seen that when the orientation $\bm{R}$ is shared (second row) \methodName{} can more precisely estimate the layouts, with improvements on all metrics relative to the single room case (first row). With a common floor and ceiling height ($d_1, d_2$) we get even better 3D and pixel-wise metrics (third row).

The choice of encoder is justified in the second section of \cref{tab:ablation-experiments}. Here we compare the ResNet-101 CNN from \cite{hanning2025pixcuboid} (third row) with our proposed \backbone{} network (last row). Despite the networks having roughly the same number of parameters (30M vs 28.5M) the \backbone{}-based one is superior, with large gains across every metric.
To isolate the impact of the learned features we run \methodName{} with just the featuremetric cost $E_{feat}$ (first two rows). \backbone{} is the best also in this case.

\begin{table*}
\centering
\caption{\textbf{Ablation experiments} on our Aria Synthetic Environments test set.}
\label{tab:ablation-experiments}
\begin{tabular}{c l c c c c c c}
  \toprule
  && \multicolumn{2}{c}{3D} & \multicolumn{2}{c}{Recall} & \multicolumn{2}{c}{Pixel-wise} \\
  \cmidrule(lr){3-4} \cmidrule(lr){5-6} \cmidrule(lr){7-8}
  && {\scriptsize IoU $\uparrow$} & {\scriptsize Chamfer $\downarrow$} & \scriptsize{Wall $\uparrow$} & \scriptsize{Room $\uparrow$} & \scriptsize{Depth $\downarrow$} & \scriptsize{Normal $\uparrow$} \\
  \midrule
  \multirow{3}{*}{\rotatebox{90}{\fontsize{7}{9}\selectfont Sharing}}
  & None & 93.5 & 0.13 m & 88.5 & 78.5 & 0.11 m & 97.9 \\
  & Orientation & 93.8 & 0.13 m & 88.9 & \textbf{79.6} & 0.10 m & 98.0 \\
  & Orientation, floor, ceiling & \textbf{94.3} & \textbf{0.12 m} & \textbf{89.0} & 79.0 & \textbf{0.09 m} & \textbf{98.2} \\
  \midrule
  \multirow{4}{*}{\rotatebox{90}{Network}}
  & ResNet-101 {\footnotesize ($E_{feat}$)} & 24.6 & 3.55 m & 8.5 & 0.5 & 1.70 m & 28.5 \\
  & \backbone{} ViT-S/14 {\footnotesize ($E_{feat}$)} & 45.1 & 2.05 m & 24.1 & 7.3 & 1.05 m & 58.1 \\
  & ResNet-101 & 82.7 & 0.43 m & 66.8 & 43.2 & 0.29 m & 93.9 \\
  & \backbone{} ViT-S/14 & \textbf{94.3} & \textbf{0.12 m} & \textbf{89.0} & \textbf{79.0} & \textbf{0.09 m} & \textbf{98.2} \\
  \midrule
  \multirow{2}{*}{\rotatebox{90}{Cost}}
  & $E_{feat}$, $E_{edge}$, $E_{VP}$ & 93.5 & 0.19 m & \textbf{89.3} & \textbf{79.1} & 0.11 m & 98.1 \\
  & $E_{feat}$, $E_{edge}$, $E_{VP}$, $E_{per}$ & \textbf{94.3} & \textbf{0.12 m} & 89.0 & 79.0 & \textbf{0.09 m} & \textbf{98.2} \\
  \midrule
  \multirow{3}{*}{\rotatebox{90}{Init.}}
  & Cuboid & 90.5 & 0.22 m & 83.4 & 73.7 & 0.16 m & 96.9 \\
  & Circle & 93.2 & 0.15 m & 87.3 & 77.7 & 0.11 m & 97.8 \\
  & Concave hull & \textbf{94.3} & \textbf{0.12 m} & \textbf{89.0} & \textbf{79.0} & \textbf{0.09 m} & \textbf{98.2} \\
  \midrule
  \multirow{3}{*}{\rotatebox{90}{Shape}}
  & No split & 93.6 & 0.14 m & \textbf{89.2} & \textbf{82.2} & 0.10 m & 98.0 \\
  & No simplification & 92.9 & 0.16 m & 86.9 & 72.5 & 0.11 m & 97.5 \\
  & Simplify \& split & \textbf{94.3} & \textbf{0.12 m} & 89.0 & 79.0 & \textbf{0.09 m} & \textbf{98.2} \\
  %\midrule
  %\multirow{2}{*}{\rotatebox{90}{\fontsize{7}{9}\selectfont Train}}
  %& Cuboid &  \\
  %& Cuboid + non-cuboid & 94.2 & 0.12 m & 88.8 & 79.3 & 0.10 m & 98.1 \\
  \bottomrule
\end{tabular}
\end{table*}

Next, the third section of \cref{tab:ablation-experiments} shows that penalizing the perimeter of the layout polygon $\mathcal{P}$ with the perimeter cost $E_{per}$ results in better 3D and pixel-wise metrics but slightly lower recall.

Three different ways to initialize the layout are compared (\cref{tab:ablation-experiments}, section four). In addition to the method shown in \cref{fig:polygon} we initialize the polygon as a cuboid (first row) and an approximate circle (second row). The cuboid is initialized following \cite{hanning2025pixcuboid} but the walls are split after the coarse and medium scales (same as for the polygons). To create the circle we compute the centroid of the camera centers in the local $xy$ plane and set the radius to be maximum distance between it and the cameras, plus a margin of 1 m. Points (4 on each quadrant) are then sampled equiangularly on the circle defined by the centroid and radius, and pairs of walls added to connect adjacent points. Both of these methods result in inferior layout predictions compared to the concave hull initialization.

Lastly, we try disabling the wall splitting and simplification (\cref{tab:ablation-experiments}, bottom section). With the proposed concave hull initialization the layout polygon typically starts with a fairly large number of walls, so \methodName{} is nearly as good without the splitting (first row compared to last).
Turning off the simplification (second row) results in overly complex layouts and a bigger performance hit.
%Turning off the simplification of the polygon (second row) is also just marginally worse, partly due to the repairing of self-intersections which can also remove walls.
% We also run our method without both splitting and simplification (third row) 

% Lastly, a model is trained exclusively on the cuboid rooms from \cite{hanning2025pixcuboid} (\cref{tab:ablation-experiments}, last section). It performs worse than training with our proposed dataset with a mix of cuboid and non-cuboid room layouts, although the differences are minor.

%In the supplementary material we show additional results, visualizations, and experiments combining our method with feed-forward pose estimators.

\subsection{Layout Estimation without Known Poses}

Our method assumes known camera poses but we believe this is a minor limitation in practice as they can easily be recovered with off-the-shelf methods such as Structure-from-Motion or recent regression-based alternatives.

As an experiment, we run $\pi^3$ \cite{wang2025pi} for each room in our ASE test set, then align the predicted poses with the ground truth using COLMAP's model aligner (to allow for evaluation) and finally estimate the layouts with \methodName{}. Results in \cref{tab:pi3-experiment} show that accuracy is only slightly worse with $\pi^3$'s poses.

\begin{table*}
\centering
\caption{\textbf{Room layout estimation} with predicted camera poses.}
\label{tab:pi3-experiment}
\begin{tabular}{l c c c c c c}
  \toprule
  & \multicolumn{2}{c}{3D} & \multicolumn{2}{c}{Recall} & \multicolumn{2}{c}{Pixel-wise} \\
  \cmidrule(lr){2-3} \cmidrule(lr){4-5} \cmidrule(lr){6-7}
  Poses & {\scriptsize IoU $\uparrow$} & {\scriptsize Chamfer $\downarrow$} & \scriptsize{Wall $\uparrow$} & \scriptsize{Room $\uparrow$} & \scriptsize{Depth $\downarrow$} & \scriptsize{Normal $\uparrow$} \\
  \midrule
  $\pi^3$ \cite{wang2025pi} & 89.4 & 0.21 m & 84.4 & 73.2 & 0.15 m & 97.0 \\
  % COLMAP \cite{schoenberger2016sfm,schoenberger2016mvs} ~~ \\
  Ground truth ~~ & \textbf{94.3} & \textbf{0.12 m} & \textbf{89.0} & \textbf{79.0} & \textbf{0.09 m} & \textbf{98.2} \\
  \bottomrule
\end{tabular}
\end{table*}

\subsection{Comparison to Robust Model Fitting on Point Clouds}
\label{subsec:comparison-to-simple-baselines}

Finally we compare against traditional baselines which try to directly fit parametric models to a 3D point cloud.
As before, we use COLMAP to produce dense point clouds from fixed poses.
To simplify the model fitting we only consider the cuboid rooms from the ScanNet++ v2 dataset.

%We additionally compare our featuremetric approach with two point-based baselines. Here cuboids are used to simplify the setup and evaluation is done on the ScanNet++ v2 dataset with cuboid rooms from \cite{hanning2025pixcuboid}.
%Again we run a dense COLMAP reconstruction for each set of images with fixed camera poses.

The first baseline method fits a cuboid directly to the dense point cloud using RANSAC~\cite{fischler1981random}, followed by non-linear refinement on the inlier set.
%The cuboid is then refined by minimizing the distance between its faces and the inlier points with L-BFGS \cite{liu1989limited}. We tune the RANSAC threshold on the validation set.
As a second baseline we run our LM optimization as usual, but replace the cost function with
$E_{dist}(\mathcal{C}) = \sum_i \rho \left( D(\mathcal{C}, \bm{P_i}) \right)$
measuring the distance $D$ between the cuboid $\mathcal{C}$ and the point cloud $\{ \bm{P}_i \}$, in an ICP-style alignment. We apply a robust loss to handle points that are not on the walls, floor or ceiling. 
For both baselines we tune parameters on the validation set.
We compare with \methodName{} using the featuremetric cost $E_{feat}$, with and without the edge cost $E_{edge}$. 
The second baseline and our method initialize the cuboids as in \cite{hanning2025pixcuboid}, but do not refine the orientation with $E_{VP}$. More details can be found in our supplementary material.

We present the results in \cref{tab:point-based-baselines}. Both baselines struggle with outlier points and cannot match our featuremetric alignment. With RANSAC the probability to sample an inlier set is low, leading to poorly fitting cuboids. The distance-based optimization performs better but is still inferior to the learned costs.

\begin{table*}
%\scriptsize
\centering
\caption{\textbf{Comparison to point-based cuboid fitting on ScanNet++ v2}. }%Results on the ScanNet++ v2 test set with cuboid rooms from \cite{hanning2025pixcuboid}.}
\label{tab:point-based-baselines}
\begin{tabular}{l c c c c c c}
  \toprule
  & \multicolumn{2}{c}{3D} & \multicolumn{2}{c}{Recall} & \multicolumn{2}{c}{Pixel-wise} \\
  \cmidrule(lr){2-3} \cmidrule(lr){4-5} \cmidrule(lr){6-7}
  & \scriptsize{IoU $\uparrow$} & \scriptsize{Chamfer $\downarrow$} & \scriptsize{Wall $\uparrow$} & \scriptsize{Room $\uparrow$} & \scriptsize{Depth $\downarrow$} & \scriptsize{Normal $\uparrow$} \\
  \midrule
  RANSAC & 26.1 & 1.34 m & 2.5 & 0.3 & 0.59 m & 36.9 \\
  $E_{dist}$ & 45.9 & 1.17 m & 19.1 & 5.0 & 0.57 m & 49.6 \\
  $E_{feat}$ & 63.5 & 0.74 m & 54.4 & 30.3 & 0.40 m & 64.2 \\
  %\hdashline
  $E_{feat} + E_{edge}$ & \textbf{92.4} & \textbf{0.15 m} & \textbf{95.0} & \textbf{88.1} & \textbf{0.06 m} & \textbf{97.2} \\
  \bottomrule
\end{tabular}
\end{table*}

\section{Conclusion}
\label{sec:conclusion}

We have introduced \methodName{}, an optimization-based method for estimating Manhattan room layouts from perspective images. Our method supports an arbitrary number of input views and can jointly optimize the layouts of multiple rooms within a building. Results on both synthetic and real datasets show that it outperforms competing methods by a large margin.

\vspace{0.5em}
{\small 
\noindent\textbf{Acknowledgments}
The work was supported by ELLIIT, the Swedish Research Council (Grant No. 2023-05424), and the Wallenberg AI, Autonomous Systems and Software Program (WASP) funded by the Knut and Alice Wallenberg Foundation. Compute was provided by the supercomputing resource Berzelius provided by National Supercomputer Centre at Linköping University and the Knut and Alice Wallenberg foundation.
The authors would like to thank Almer Hanning for annotating the ScanNet++ room layouts.
}

\newpage
{
\small This version of the manuscript has been accepted for publication, after final editorial and peer review (where applicable) is complete, but is not the Version of Record and does not reflect post-acceptance improvements (such as copyediting or typesetting), or any corrections. The final authenticated version is available online at: \url{https://dx.doi.org/10.1007/978-3-032-37023-5_18}. Use of this Accepted Manuscript version is subject to the publisher’s Accepted Manuscript terms of use: \url{https://www.springernature.com/gp/open-research/policies/accepted-manuscript-terms}
}

}

%\clearpage  % TODO FINAL: This \clearpage needs to be removed from both review and camera-ready versions.

%\section*{Acknowledgements}
%Please insert your acknowledgments here.

\bibliographystyle{splncs04}
\bibliography{main}

% For arXiv version, set doSupplementary to false and uncomment the following line

\end{document}